%% file: main.tex
\documentclass[10pt,twocolumn,letterpaper]{article}

\usepackage[pagenumbers]{cvpr} %

\input{preamble}
\input{math_commands.tex}

\definecolor{cvprblue}{rgb}{0.21,0.49,0.74}
\usepackage[pagebackref,breaklinks,colorlinks,allcolors=cvprblue]{hyperref}

\usepackage{url}            %
\usepackage{nicefrac}       %
\usepackage{microtype}      %
\usepackage{graphicx}
\usepackage{amsmath}
\usepackage{amssymb}
\usepackage{booktabs}
\usepackage{xspace}
\usepackage{amsmath,amsfonts,amssymb,amsthm}
\usepackage[dvipsnames]{xcolor}
\usepackage{diagbox}
\usepackage{pifont}
\usepackage{svg}
\usepackage{enumitem}
\usepackage{physics}
\usepackage{mathtools}
\usepackage{algorithm}
\usepackage{algorithmic}
\usepackage{gensymb}
\usepackage{wrapfig}
\usepackage{subcaption}
\usepackage{multirow}
\usepackage{footmisc}
\usepackage{empheq}
\usepackage{cases}
\usepackage{adjustbox}
\usepackage{array}
\usepackage{tabu}
\usepackage{flushend}
\usepackage{multirow}
\usepackage{bm}
\usepackage{color, colortbl}
\usepackage{makecell}
\usepackage{capt-of,etoolbox}

\def\paperID{xxxx} %
\def\confName{CVPR}
\def\confYear{2026}

\title{Attend Before Attention: \\ Efficient and Scalable Video Understanding via Autoregressive Gazing}

\author{Baifeng Shi\textsuperscript{1,4*} \qquad Stephanie Fu\textsuperscript{1*} \qquad Long Lian\textsuperscript{1} \qquad Hanrong Ye\textsuperscript{4} \\[3pt]  David Eigen\textsuperscript{3} \qquad Aaron Reite\textsuperscript{3}  \qquad Boyi Li\textsuperscript{1,4} \qquad Jan Kautz\textsuperscript{4} \qquad Song Han\textsuperscript{2,4} \\[3pt] David M. Chan\textsuperscript{1\textdagger} \qquad Pavlo Molchanov\textsuperscript{4\textdagger} \qquad Trevor Darrell\textsuperscript{1\textdagger} \qquad Hongxu Yin\textsuperscript{4\textdagger} 
\\[12pt] \textsuperscript{1}UC Berkeley \quad \textsuperscript{2}MIT \quad \textsuperscript{3}Clarifai \quad \textsuperscript{4}NVIDIA
}

\makeatletter
\apptocmd\@maketitle{{\myfigure{}\par}}{}{}
\makeatother

\begin{document}

\makeatletter
\DeclareRobustCommand\onedot{\futurelet\@let@token\@onedot}
\def\@onedot{\ifx\@let@token.\else.\null\fi\xspace}

\def\eg{e.g.\xspace} \def\Eg{E.g\onedot}
\def\ie{i.e.\xspace} \def\Ie{I.e\onedot}
\def\vs{vs.\xspace}
\def\cf{c.f\onedot} \def\Cf{C.f\onedot}
\def\etc{etc\onedot} \def\vs{vs\onedot}
\def\wrt{w.r.t\onedot} \def\dof{d.o.f\onedot}
\def\etal{et al\onedot}
\def\viz{viz\onedot}
\def\aka{a.k.a\onedot}

\def\CN{\textcolor{blue}{CIT. Needed}}

\def\blfootnote{\xdef\@thefnmark{}\@footnotetext}

\makeatother

\newcommand{\customfootnotetext}[2]{{%
  \renewcommand{\thefootnote}{#1}%
  \footnotetext[0]{#2}}}%

\newcommand{\model}{AutoGaze\xspace}
\newcommand{\stwo}{{S$^2$}\xspace}
\newcommand{\stwobf}{{\textbf{S$\mathbf{^2}$}}\xspace}
\newcommand{\stwowrapper}{{S$^2$-Wrapper}\xspace}
\newcommand{\benchmark}{HLVid\xspace}

\definecolor{darkgreen}{HTML}{054907}
\definecolor{darkgreen_teaser}{HTML}{105611}
\definecolor{lightgreen}{HTML}{9CCBB8}
\definecolor{lightred}{HTML}{E3242B}
\definecolor{lightorange}{HTML}{ED7D31}
\definecolor{darkorange}{HTML}{D8976F}
\definecolor{nvidiagreen}{HTML}{76B900}

\newif\ifcomments
\commentstrue        %

\newcommand{\cmark}{\textcolor{OliveGreen}{\ding{51}}}%
\newcommand{\xmark}{\textcolor{Maroon}{\ding{55}}}%

\newcommand{\bs}[1]{\textcolor{red}{\textbf{[Baifeng: #1]}}}
\newcommand{\stephanie}[1]{\textcolor{PineGreen}{\textbf{[Stephanie: #1]}}}
\newcommand{\tony}[1]{\textcolor{blue}{\textbf{[Long (Tony): #1]}}}
\newcommand{\yin}[1]{\textcolor{cyan}{\textbf{[Yin: #1]}}}
\newcommand{\PM}[1]{\textcolor{orange}{\textbf{[Pavlo: #1]}}}
\newcommand{\ye}[1]{\textcolor{green}{\textbf{[YE: #1]}}}

\ifcomments\else
  \renewcommand{\bs}[1]{}
  \renewcommand{\stephanie}[1]{}
  \renewcommand{\tony}[1]{}
  \renewcommand{\yin}[1]{}
  \renewcommand{\PM}[1]{}
  \renewcommand{\ye}[1]{}
\fi

\newcommand\minisection[1]{\vspace{1.3mm}\noindent \textbf{#1}}

\newcommand{\drawio}[1]{{\color{Gray}{\bf drawio: #1}}}
\newcommand{\han}[1]{{\color{blue}{\bf Han: #1}}}

\newcommand{\PreserveBackslash}[1]{\let\temp=\\#1\let\\=\temp}
\newcolumntype{C}[1]{>{\PreserveBackslash\centering}p{#1}}
\newcolumntype{R}[1]{>{\PreserveBackslash\raggedleft}p{#1}}
\newcolumntype{L}[1]{>{\PreserveBackslash\raggedright}p{#1}}

\newcommand{\ver}[1]{\rotatebox[origin=l]{90}{#1}}
\newcommand{\multirowver}[1]{\rotatebox[origin=l]{90}{\parbox{1.5cm}{#1}}}

\renewcommand{\minisection}[1]{\noindent\textbf{#1}}

\setlength{\textfloatsep}{0pt}
\setlength{\textfloatsep}{20pt plus 2pt minus 4pt}
\setlength{\textfloatsep}{10pt plus 2pt minus 4pt}
\setlength{\textfloatsep}{10pt plus 1pt minus 2pt}
\setlength{\dbltextfloatsep}{3pt}
\setlength{\abovecaptionskip}{3pt}
\setlength{\belowcaptionskip}{3pt}

\renewcommand{\topfraction}{0.99}        %
\renewcommand{\bottomfraction}{0.99}     %
\renewcommand{\textfraction}{0.01}       %
\renewcommand{\floatpagefraction}{0.99}  %

\hypersetup{
    colorlinks=true,
    linkcolor=lightorange,      %
    citecolor=lightorange,       %
    urlcolor=lightorange     %
}

\newcommand\myfigure{%
\centering
\vspace{-3mm}
    \includegraphics[width=\linewidth]{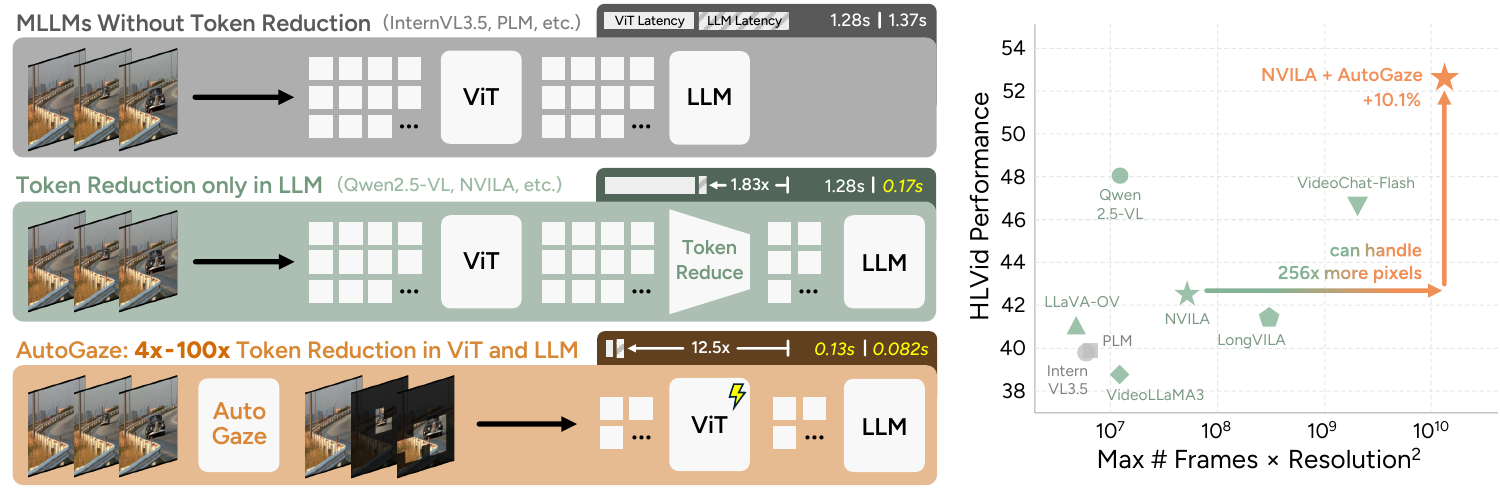}
\vspace{-4mm}
\captionof{figure}{We propose \textbf{\model}, which \textbf{reduces the computational cost of video understanding} to  \textbf{scale MLLMs to long, high-resolution videos}. \textit{(Left)} Existing MLLMs either process all pixels which is inefficient, or prune tokens only in their LLMs, leaving ViTs the computational bottleneck. In contrast, \model eliminates redundant patches by up to 100$\times$ \emph{before} ViTs, accelerating ViTs and MLLMs by up to 19$\times$. \textit{(Right)} This efficiency enables MLLMs with \model to scale to 1K-frame, 4K-resolution videos and achieve superior performance on \benchmark, our new long, high-resolution video benchmark, surpassing prior MLLMs limited to short or low-resolution videos.
}

\label{fig:teaser}
\vspace{1.4em}
}

\maketitle

\customfootnotetext{ }{\textsuperscript{*}Equal contribution. \ \textsuperscript{\textdagger}Equal advising.}

\input{sections/0_abstract}
\input{sections/1_intro}

\input{sections/2_related_work}

\input{sections/3_building_autogaze}
\input{sections/4_exp}

\input{sections/5_discussion}

{
    \small
    \bibliographystyle{ieeenat_fullname}
    \bibliography{main}
}

\clearpage
\appendix
\setcounter{page}{1}

\input{sections/appendix}

\end{document}

%% file: math_commands.tex
\usepackage{amsmath,amsfonts,bm}

\def\eqref#1{equation~\ref{#1}}

\def\1{\bm{1}}

\def\vs{{\bm{s}}}

\def\mX{{\bm{X}}}

\DeclareMathAlphabet{\mathsfit}{\encodingdefault}{\sfdefault}{m}{sl}
\SetMathAlphabet{\mathsfit}{bold}{\encodingdefault}{\sfdefault}{bx}{n}

%% file: sections/0_abstract.tex
\begin{abstract} 
Multi-modal large language models (MLLMs) have advanced general-purpose video understanding but struggle with long, high-resolution videos---they process every pixel equally in their vision transformers (ViTs) or LLMs despite significant spatiotemporal redundancy.
We introduce \textbf{\model}, a lightweight module that removes redundant patches before processed by a ViT or an MLLM. 
Trained with next-token prediction and reinforcement learning,
\model autoregressively selects a minimal set of multi-scale patches that can reconstruct the video within a user-specified error threshold, eliminating redundancy while preserving information.
Empirically, \model reduces visual tokens by 4$\times$-100$\times$ and 
accelerates ViTs and MLLMs 
by up to 19$\times$, enabling scaling MLLMs to 1K-frame 4K-resolution videos and achieving superior results on video benchmarks (e.g., 67.0\% on VideoMME).
Furthermore, we introduce \textbf{\benchmark}: the first high-resolution, long-form video QA benchmark with 5-minute 4K-resolution videos, where an MLLM scaled with \model improves over the baseline by 10.1\% and outperforms the previous best MLLM by 4.5\%. Project page: \url{https://autogaze.github.io/}.

\end{abstract}

\vspace{-1em}

%% file: sections/1_intro.tex
\section{Introduction}

\begin{figure*}
  \begin{center}
    \includegraphics[width=1.0\textwidth]{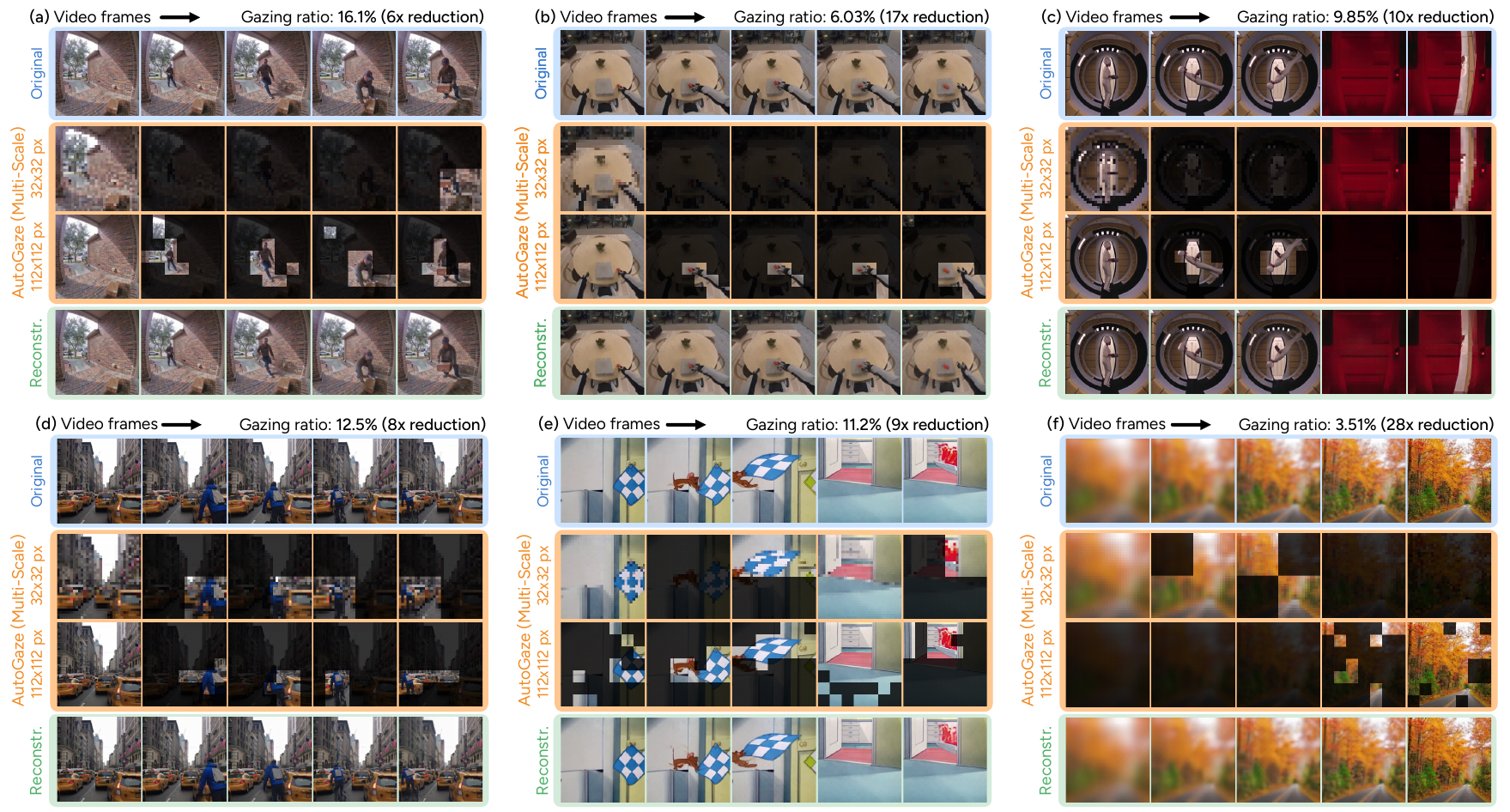}
  \end{center}

  \vspace{-0.8em}
  \caption{
    \small \textbf{What is \model paying attention to?} For each example, we show the original video, multi-scale gazed patches, and reconstructed video. Note that we only show gazing on two scales to save space while \model actually uses four. In general, \model can \textbf{1)} focus on moving objects while removing redundancy in static regions (a-e), \textbf{2)} adapt to scene changes by selecting more patches (c, e), and \textbf{3)} distribute attention with different granularity based on detailedness (c, f). This allows \model to select a small ratio of patches (gazing ratio) without much information loss, as reflected by the reconstruction quality. 
    }
    \label{fig:gazing_visualization}
\end{figure*}

When observing a moving scene, humans don't process every detail equally. Our eyes dart around to moving objects, capture fine details, and skip over static backgrounds, 
efficiently understanding scenes by selectively attending to informative regions~\cite{itti2001computational,itti2002model,rensink2000dynamic,anderson1995cognitive}.
This allows us to process high-FPS, high-resolution video streams in real time.
In contrast, modern video understanding models (e.g., multi-modal large language models (MLLMs)~\cite{hurst2024gpt,team2024gemini,li2024llava,liu2025nvila,bai2025qwen2}) still process every pixel in every frame equally, wasting computation due to spatiotemporal redundancy in videos~\cite{le1991mpeg,wiegand2003overview,sullivan2012overview,wu2018video,chen2021nerv}. 
For example, in Fig. \ref{fig:gazing_visualization} (top-left), the static background only needs to be viewed once. Thus, these models cannot scale to \textit{long-form} and \textit{high-resolution} videos crucial for real-world applications~\cite{gao2025magicdrive,ren2025vista,deliege2021soccernet,corona2021meva,ye2025omnivinci} due to computational cost.

Recent work attempts to reduce video redundancy in MLLMs, but typically prunes tokens only in the LLM while the vision transformer (ViT) still processes all pixels, creating a huge efficiency bottleneck that prevents scaling to longer, higher-resolution videos~\cite{shen2024longvu,ye2024mm,li2024videochat,shen2025fastvid,tang2025adaptive} (Fig. \ref{fig:teaser}). Moreover, these methods either rely on heuristics such as attention scores which underperforms learned approaches~\cite{shi2025scaling} or involves heavy search and reasoning that adds overhead and further limits scalability~\cite{ye2025re,wang2025videoitg,wang2024videoagent,yu2024frame}.

To this end, we propose \textbf{\model}, a 3M-parameter lightweight model that attends to informative patches and removes redundant ones \textit{before} a ViT. Specifically, \model perceives each frame and autoregressively selects a minimal set of \textit{multi-scale} patches which, along with the selected patches from previous frames, can reconstruct the current frame within a user-specified reconstruction loss threshold. This model, pre-trained with next-token-prediction on a curated dataset of gazing sequences and post-trained with RL on reconstruction rewards, learns to focus only on newly emerged content while ignoring repeated information, and use multi-scale patches to cover broad areas coarsely, zooming in on fine details where needed. For example, Fig. \ref{fig:gazing_visualization} shows \model removing redundant patches in static regions and selecting coarser scales in low-detail areas. By only processing the selected multi-scale patches, both ViTs and LLMs are substantially sped up, unlocking efficient processing of long, high-resolution videos (Fig. \ref{fig:teaser}).

Empirically, \model reduces the number of patches by 4$\times$-100$\times$ for videos with different FPS and resolution (\eg, 1\% patches for 30-FPS 4K-resolution videos) while maintaining downstream MLLM performance. This leads to up to 19$\times$ and 10$\times$ speedup for ViTs and MLLMs. Leveraging this efficiency, we scale an MLLM (NVILA~\cite{liu2025nvila}) to 1K-frame 4K-resolution videos, demonstrating consistent improvements on various benchmark (e.g., 67.0\% on VideoMME~\cite{fu2025video}) and outperforming strong MLLMs such as Qwen2.5-VL~\cite{bai2025qwen2}. We also show that \model generalizes to videos with out-of-distribution styles and semantics. 

Furthermore, noticing that existing benchmarks only focus on long videos but not high resolution~\cite{mangalam2023egoschema,wu2024longvideobench,wang2025lvbench,fu2025video,zhou2024mlvu}, we introduce \textbf{\benchmark}, the first high-resolution, long-form video QA benchmark, to stress-test \model's scalability. It consists of 268 QAs about details in up to 5-minute, 4K-resolution videos, requiring visual perception at 1K - 2K resolution to solve. We show that scaling an MLLM \cite{liu2025nvila} to 1K frames and 4K resolution via \model significantly improves its performance from 42.5\% to 52.6\%, outperforming the previous best MLLM~\cite{li2024videochat} by 4.5\% (Fig. \ref{fig:teaser}).

%% file: sections/2_related_work.tex
\vspace{-0.3em}
\section{Related Work}
\vspace{-0.3em}
\minisection{Video understanding and Long-Context MLLMs.} Classical video understanding has long been driven by supervised or self-supervised video encoders including 3D-ConvNets and early transformers \cite{carreira2017quo,feichtenhofer2019slowfast,feichtenhofer2021large,arnab2021vivit}, and pre-training algorithms such as masked auto-encoding \cite{tong2022videomae,feichtenhofer2022masked,wang2023videomae,bardes2024revisiting,assran2025v}, predictive coding~\cite{vondrick2016anticipating,han2019video,rajasegaran2025empirical}, and large-scale vision-language pre-training \cite{xu2021videoclip,wang2023internvid,yan2022videococa,bolya2025perception,wang2022internvideo,wang2024internvideo2}. 
Recent MLLMs have extended these encoders to general-purpose video QA and captioning \cite{bai2025qwen2,hurst2024gpt,li2024llava,liu2025nvila,team2024gemini,zhang2025videollama}.
However, these models usually operate on short, low-resolution clips due to costs of scaling to higher spatiotemporal resolution.
While new long-video benchmarks \cite{mangalam2023egoschema,wu2024longvideobench,zhou2024mlvu} and models~\cite{chen2024longvila,chen2025scaling,li2024videochat,zhang2024long,pang2025mr} emphasize extended temporal understanding, they remain limited to low resolutions due to inefficient whole-video processing, leaving a gap for methods and benchmarks that support \textit{both} thousand-frame context and 4K-resolution detail under realistic compute constraints. 

\minisection{Token Reduction and Compression.} A rapidly growing line of work has targeted ViT and MLLM efficiency by reducing input tokens. Spatial methods \cite{shi2025scaling,bolya2022token,yuan2024efficient,cao2023pumer,yang2025visionzip,rao2021dynamicvit,pan2021iared2,li2022sait} compress tokens or select informative patches based on attention scores or task relevance. Temporal methods reduce frame redundancy via sub-sampling \cite{wang2016tsn}, segment-level pooling \cite{fan2021mvit,ren2023testa}, or learned keyframe selection \cite{tang2025aks,zhu2025focus}; spatiotemporal schemes such as STORM \cite{jiang2025storm}, FastVID \cite{shen2025fastvid}, LongVU \cite{shen2024longvu}, and VideoChat-Flash \cite{li2024videochat}, either simply pool tokens or use the ViT features to prune or aggregate tokens. However, all of these models only prune tokens inside the model or between the ViT and LLM, leaving part of the model still processing the full video at high cost. In contrast, \model removes redundant patches \textit{before} the ViT, significantly improving efficiency. Other works on adaptive tokenization lean where to allocate tokens rather than using a fixed uniform grid~\cite{duggal2024alit,duggal2025single,yu2024titok,bachmann2025flextok,yan2024elastictok}. However, their large tokenizer adds additional computational overhead and the tokenization is not adaptive to pre-trained ViTs.

%% file: sections/3_building_autogaze.tex
\begin{figure*}[t]
  \begin{center}
    \includegraphics[width=1.0\textwidth]{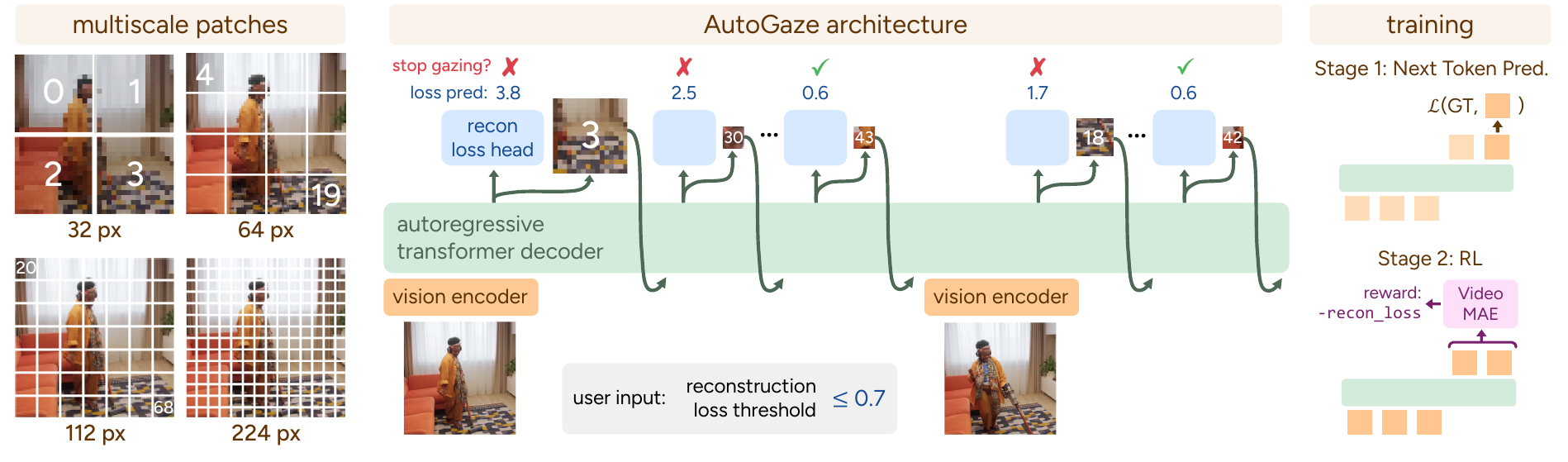}
  \end{center}

  \vspace{-1.3em}
  \caption{
    \small \textbf{Architecture and training pipeline of \model.} \textit{(Left \& Middle)} Given a video, \model processes each frame and autoregressively decodes indices of multi-scale patches based on the history of frames and selected patches. Once it believes the previously-gazed patches are sufficient to reconstruct the current frame, it automatically stops gazing and moves to the next frame. \textit{(Right)} \model is trained in two stages: next-token-prediction pre-training on collected gazing sequences, and RL post-training with reconstruction reward.
  }
    \label{fig:architecture}
\end{figure*}

\section{\model for Efficient Video Understanding}
\label{sec:building_autogaze}
\vspace{-0.3em}

Given a video, \model selects a minimal set of patches (i.e., ``gazing'') which can reconstruct the video within a reconstruction loss threshold. Formally, for a $T$-frame video $\mX^{1:T}$ where $\mX^t$ is the $t$-th frame and each frame contains $V$ patches, \model outputs a set of patch indices:
\begin{equation}\small
     \texttt{AutoGaze}: \mX^{1:T} \rightarrow p^1_{1:N^1}, \dots, p^T_{1:N^T},
\end{equation}
where $p^t_k \in \{1, \dots, V\}$ is the index of the $k$-th patch selected at frame $t$, and $N^t$ is the number of selected patches (or ``gazing length'') at frame $t$. 

To select the minimal set satisfying the threshold, \model is able to select  patches that minimize reconstruction loss under \textit{any} $N^{1:T}$ and find the smallest $N^{1:T}$ satisfying the threshold. Formally, given any $N^{1:T}$, \model can predict patch indices $p^1_{1:N^1}, \dots, p^T_{1:N^T}$ that optimize
\begin{equation}\small
\label{eq:objective}
    \min_{p^1_1, \dots, p^T_{N^T}} L(\mX^{1:T}, \  \texttt{Recon}(\mX^1[p^1_1], \dots, \mX^T[p^T_{N^T}])),
\end{equation}
where $\mX^t[p^t_k]$ is the $p^t_k$-th patch in frame $t$, $\texttt{Recon}(\cdot)$ is a model that reconstructs the original video from the gazed patches, and $L(\cdot, \cdot)$ is a distance function between the original and the reconstructed videos. We instantiate $\texttt{Recon}(\cdot)$ as a custom VideoMAE~\cite{tong2022videomae} with block-causal attention, and $L(\cdot, \cdot)$ as a weighted sum of pixel reconstruction loss and perceptual loss~\cite{zhang2018unreasonable,johnson2016perceptual} (see Appendix~\ref{appendix_sec:model_detail} for details). At the same time, \model can identify the smallest $N^{1:T}$ that satisfies $L^\ast(N^{1:T}) < \epsilon$ where $L^\ast(N^{1:T})$ is the optimal reconstruction loss under gazing lengths $N^{1:T}$ (Eq. \ref{eq:objective}) and $\epsilon$ is a user-specified loss threshold.

To achieve this, we build \model to \textbf{autoregressively select patch indices} that optimize reconstruction loss for any gazing length, while \textbf{automatically deciding the smallest gazing length} by predicting reconstruction loss on the fly and stopping once it falls below the threshold. Below, we introduce model design (Sec. \ref{sec:autogaze_model}), training pipeline (Sec. \ref{sec:autogaze_training}), how to apply it to videos of any duration and resolution, and integrate it into any ViT (Sec. \ref{sec:using_autogaze}), and a new benchmark to stress-test scalability (Sec. \ref{sec:benchmark}).

\subsection{Model Design}
\label{sec:autogaze_model}

Fig. \ref{fig:architecture} (Middle) illustrates \model's lightweight design: a convolutional encoder and autoregressive transformer decoder, totaling 3M parameters.

\minisection{Autoregressive gazing.} Given a video, \model interleaves frame encoding and patch gazing. It starts by encoding the first frame with the convolutional encoder, passing the features to the decoder, and autoregressively decoding patch indices. The decoding process mirrors LLMs except the vocabulary contains only patch indices $\{1, \dots, V\}$ instead of words. 
Next, \model encodes the second frame and decodes its patch indices based on \textit{the features of both frames and the gazed patch indices of the first frame}. This lets the model avoid redundant patches by referring to frame and gazing history. The process repeats for subsequent frames.

\minisection{Automatically deciding the gazing length.} 
To identify the smallest $N^t$ satisfying the reconstruction loss threshold, we add a head on the decoder that, when decoding every $p_k^t$, predicts the loss of reconstructing frame $t$ from the patches gazed up to that step, \ie, $\{p_1^1, \dots, p_k^t\}$. Once the predicted loss falls below the threshold, it stops gazing for that frame. 

\minisection{Multi-scale gazing.} Considering that not all regions need full resolution (\eg, solid-colored regions can be stored losslessly in low resolution), \model supports multi-scale gazing. The decoder's vocabulary includes patches from multiple scales (Fig. \ref{fig:architecture} (Left)),
letting the decoder select different scales for regions with different level of detail, reducing patches while preserving reconstruction quality (Sec. \ref{sec:exp_ablation}). This also requires the downstream ViT to accept multi-scale patches as input, which we detail in Sec. \ref{sec:using_autogaze}.

\minisection{Multi-token prediction.} We adopt multi-token prediction~\citep{gloeckle2024better} by using multiple heads to output multiple patch indices and corresponding reconstruction losses at once, speeding up gazing with little performance loss (Sec. \ref{sec:exp_ablation}).

\subsection{Training Pipeline}
\label{sec:autogaze_training}

\model is trained to decode patch indices that minimize reconstruction loss at any gazing length and predict reconstruction loss at each step for automatic stopping. Inspired by modern LLM training~\citep{radford2018improving,ouyang2022training,achiam2023gpt,guo2025deepseek}, we train \model in two stages (Fig. \ref{fig:architecture} (Right)). 
First, we pre-train with next-token prediction (NTP) on videos paired with ground-truth gazing sequences that are collected via greedy search to approximately minimize reconstruction loss.
Next, since the pre-trained gazing quality is bounded by the sub-optimal gazing data, we further post-train \model using RL with reconstruction reward to discover gazing sequences with lower reconstruction loss. We also train reconstruction loss prediction in both stages to enable automatic stopping. 

\vspace{0.5em}
\minisection{Pre-training with next-token-prediction (NTP).} 
Given a dataset with pairs of video $\mX^{1:T}$, gazing sequences $\{\widetilde{p}^1_1, \dots, \widetilde{p}^T_{N^T}\}$ that approximately minimize reconstruction loss under random gazing length $N^{1:T}$, and $\{\widetilde{l}^1_1, \dots, \widetilde{l}^T_{N^T}\}$ where $\widetilde{l}^t_k$ records reconstruction loss of frame $t$ after gazing at $\widetilde{p}^t_k$, we pre-train \model with NTP cross-entropy loss
\vspace{-0.8em}
\begin{equation}\small
\label{eq:ntp_loss}
    L_{NTP} = - \sum_{t=1}^T \sum_{k=1}^{N^t} \log \pi_\theta(\widetilde{p}_k^t \mid \mX^{1:t}, \widetilde{p}_{1:N^1}^1, \dots, \widetilde{p}_{1:k-1}^t),
\end{equation}
where $\pi_\theta$ is the model and $\pi_\theta(\widetilde{p}_k^t \mid \mX^{1:t}, \widetilde{p}_{1:N^1}^1, \dots, \widetilde{p}_{1:k-1}^t)$ is the probability of decoding $\widetilde{p}_k^t$ based on previous frames and gazing. We also supervise reconstruction loss prediction with an $\ell_2$ loss using $\{\widetilde{l}^1_1, \dots, \widetilde{l}^T_{N^T}\}$. \model thus learns sub-optimal gazing at different gazing length and learns to predict reconstruction loss at each decoding step.

\minisection{Post-training with RL.} 
Since the pre-training data only contains sub-optimal gazing, we further improve \model with RL post-training, using a simplified, on-policy GRPO \citep{shao2024deepseekmath,liu2025understanding} algorithm with reconstruction loss as reward:
\begin{equation}\small
    L_{GRPO} = -\sum_{t=1}^T \sum_{k=1}^{N^t} \frac{\pi_\theta(p_k^t)}{\pi_{\theta_{detached}}(p_k^t)} A_k^t,
\end{equation}
where $\pi_\theta(p_k^t)$ is short for the decoding probability of patch index $p_k^t$ as in Eq. \ref{eq:ntp_loss}, $\pi_{\theta_{detached}}$ is $\pi_\theta$ without gradient, and advantage $A_k^t$ is the return $G_k^t$ normalized within the group of GRPO where $G_k^t = \sum_{\tau=t}^T \gamma^{N^t - k + \sum_{s=t+1}^\tau N^s} \cdot (-l_{N^\tau}^\tau)$, i.e., sum of negative reconstruction loss of future frames discounted by $\gamma$. Additionally, we supervise reconstruction loss prediction at the last patch of each frame (\ie, $l_{N^t}^t$) using the actual reconstruction loss at frame $t$.

\minisection{Training data curation.} The training pipeline above requires raw videos and paired gazing sequences for pre-training. We first collect a set of 800K videos spanning egocentric, exocentric, natural, and text-rich videos. Each video is sampled at 16 frames and 224 resolution. We then collect gazing sequences that approximately minimize reconstruction loss for 250K videos using greedy search. Specifically, we start from the first patch of the first frame and exhaustively find which patch gives the lowest reconstruction loss. We repeat this until reaching the first frame's gazing length, then proceed to the second frame and so on. We also record reconstruction loss at each step to supervise loss prediction. See Appendix~\ref{appendix_sec:training_detail} for details.

\subsection{Downstream Usage of \model}
\label{sec:using_autogaze}

\minisection{Inference on videos with any resolution and duration.} Despite being trained on 16-frame 224$\times$224 videos, \model processes videos of any resolution and duration without additional training. Inspired by any-resolution MLLMs~\citep{chen2024far,liu2024llavanext,shi2024we}, we split the video into 16$\times$224$\times$224 spatiotemporal tiles, run \model on each tile, and merge the gazed positions back together, allowing \model to scale to 1K-frame and 4K-resolution videos (Sec. \ref{sec:exp}).

\minisection{Integrating \model into ViTs and MLLMs.} Current MLLMs typically encode each full frame using an image ViT~\cite{bai2025qwen2,wang2025internvl3,liu2025nvila}. To integrate \model, we make two changes. First, we allow ViTs to take multi-scale patch input by interpolating each frame and positional embeddings to different scales, running patch embedding on each scale separately, and then feeding embedded tokens from all scales to the ViT. Second, we repurpose image ViTs into video ViTs by letting them process tokens from all 16 frames in the same sequence. With these changes, \model selects multi-scale patches for a video, encodes them with a ViT, and the encoded tokens can be fed into MLLMs as usual.

\subsection{\benchmark: A High-Res, Long Video Benchmark}
\label{sec:benchmark}

Although \model enables efficient understanding of long, high-resolution videos, benchmarks to evaluate this capability are still missing---current benchmarks~\cite{wu2024longvideobench,mangalam2023egoschema,song2024moviechat} only focus on long videos with several minutes of duration but not high resolution. 
To this end, we propose \benchmark, the first long-form, high-resolution video QA benchmark featuring 268 QA pairs on up to 5-minute, 4K-resolution videos. 
Each question is manually reviewed to ensure high resolution is required. Details are deferred to Appendix~\ref{appendix_sec:benchmark_detail}, and some examples from the benchmark are visualized in Fig. \ref{fig:hlvid_examples}. We find that an MLLM scaled to 1K frames and 4K resolution via \model achieves significant improvement and unlocks state-of-the-art performance on \benchmark (Sec. \ref{sec:exp_performance}).

%% file: sections/4_exp.tex
\section{Experiments}
\label{sec:exp}

\begin{figure}
  \begin{center}
    \includegraphics[width=1.0\linewidth]{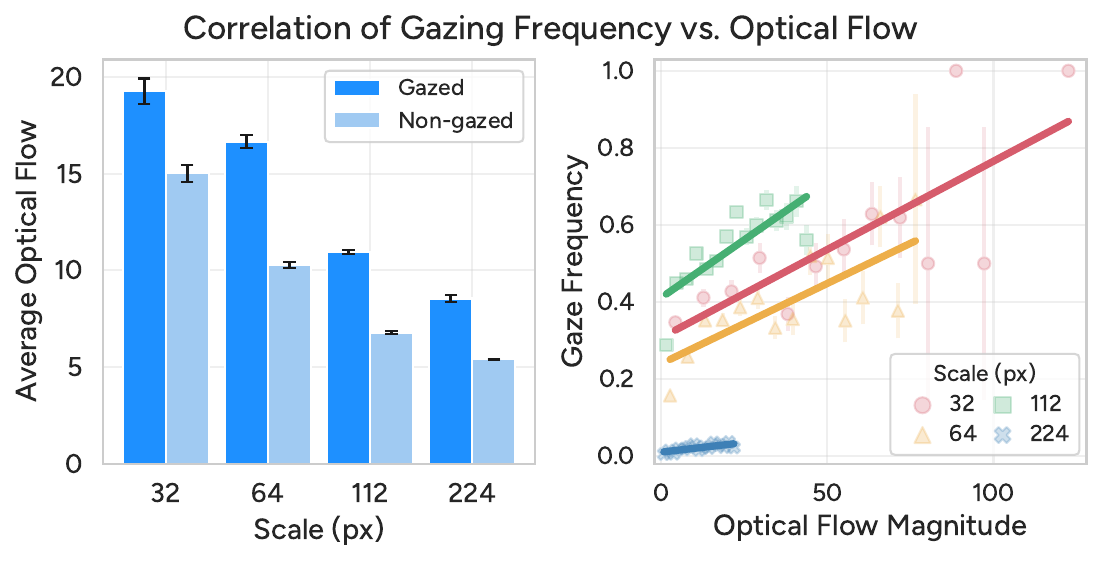}
  \end{center}
    \vspace{-1em}
  \caption{\textbf{\model targets patches with higher optical flow.} \textit{(Left)} \model uses coarser scales to capture higher optical flow.
  \textit{(Right)} Across all scales, \model more frequently selects patches with higher optical flow. 
  Error bars represent SEM.
  }
  \label{fig:gazing_flow}
  \vspace{-0.5em}
\end{figure}

\begin{figure}
  \begin{center}
    \includegraphics[width=1.0\linewidth]{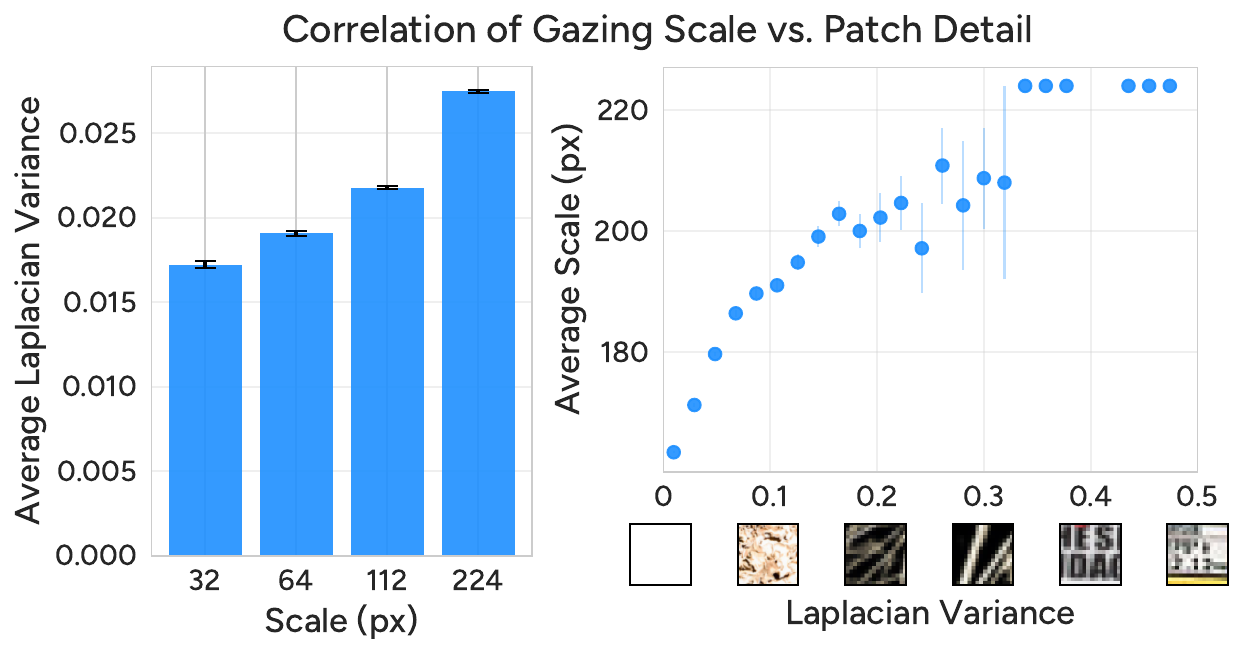}
  \end{center}
    \vspace{-1em}
  \caption{\textbf{Gazing scale correlates with patch detail.} \textit{(Left)} At finer scales, \model selects more detailed patches (measured as Laplacian variance). \textit{(Right)} With increasing detail, \model uses finer scales ($\rho = .12, p < 0.001$). Sample patches with Laplacian variances are shown below the x-axis. Error bars represent SEM.}
  \label{fig:gazing_texture}
  \vspace{-0.5em}
\end{figure}

\begin{figure*}
  \begin{center}
    \includegraphics[width=1.0\linewidth]{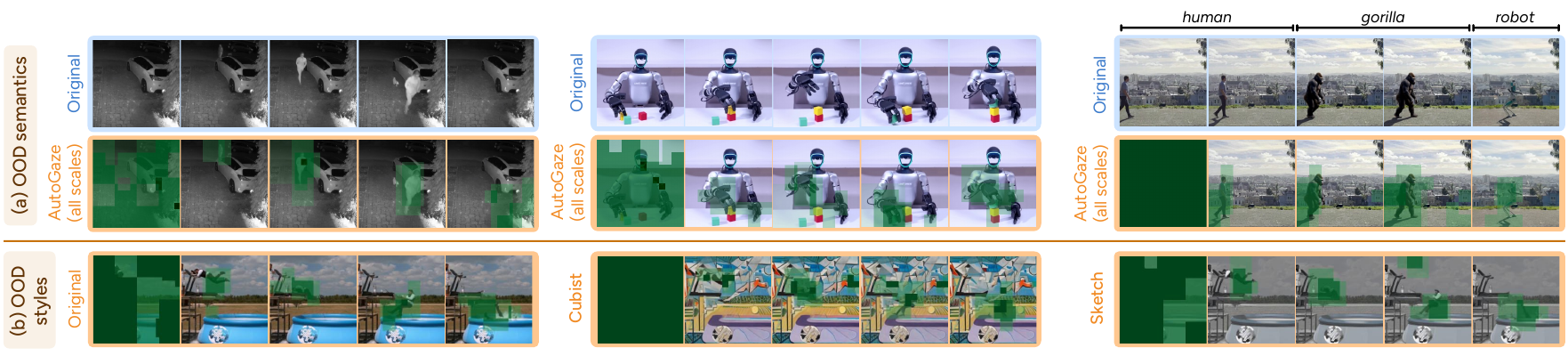}
   \vspace{-1.6em}
  \end{center}
  \caption{\textbf{Generlizability of \model to OOD videos.} 
  \textit{(a)} We show model behavior on videos with OOD semantics, including a CCTV clip (left), robot grasping demo (middle), and a video with object swapping (right). In each example, \model still robustly tracks the changing parts despite the unseen semantics, object categories, and unexpected changes.
  \textit{(b)} We show \model output on the same video with different style transfer. \model consistently tracks the falling person regardless of visual style, texture and global illumination. 
  }
  \label{fig:ood}
\end{figure*}

We evaluate \model’s behavior, efficiency, and performance. Sec.~\ref{sec:exp_analysis} examines which patches \model selects or ignores and tests its generalization to unseen video styles and semantics. Sec.~\ref{sec:exp_efficiency} measures its efficiency gains for ViTs and MLLMs. Leveraging the efficiency, Sec.~\ref{sec:exp_performance} shows that \model enables higher-resolution and longer video processing in MLLMs with improved performance. Sec.~\ref{sec:exp_baseline_comparison} compares \model against gazing and MLLM token-reduction baselines, and Sec.~\ref{sec:exp_ablation} ablates training and modeling choices. We use SigLIP2-SO400M~\cite{tschannen2025siglip} and NVILA-8B-Video~\cite{liu2025nvila} as the ViT and MLLM by default.

\subsection{What is \model paying attention to?}
\label{sec:exp_analysis}

\begin{figure}[t]
  \begin{center}
    \includegraphics[width=1.0\linewidth]{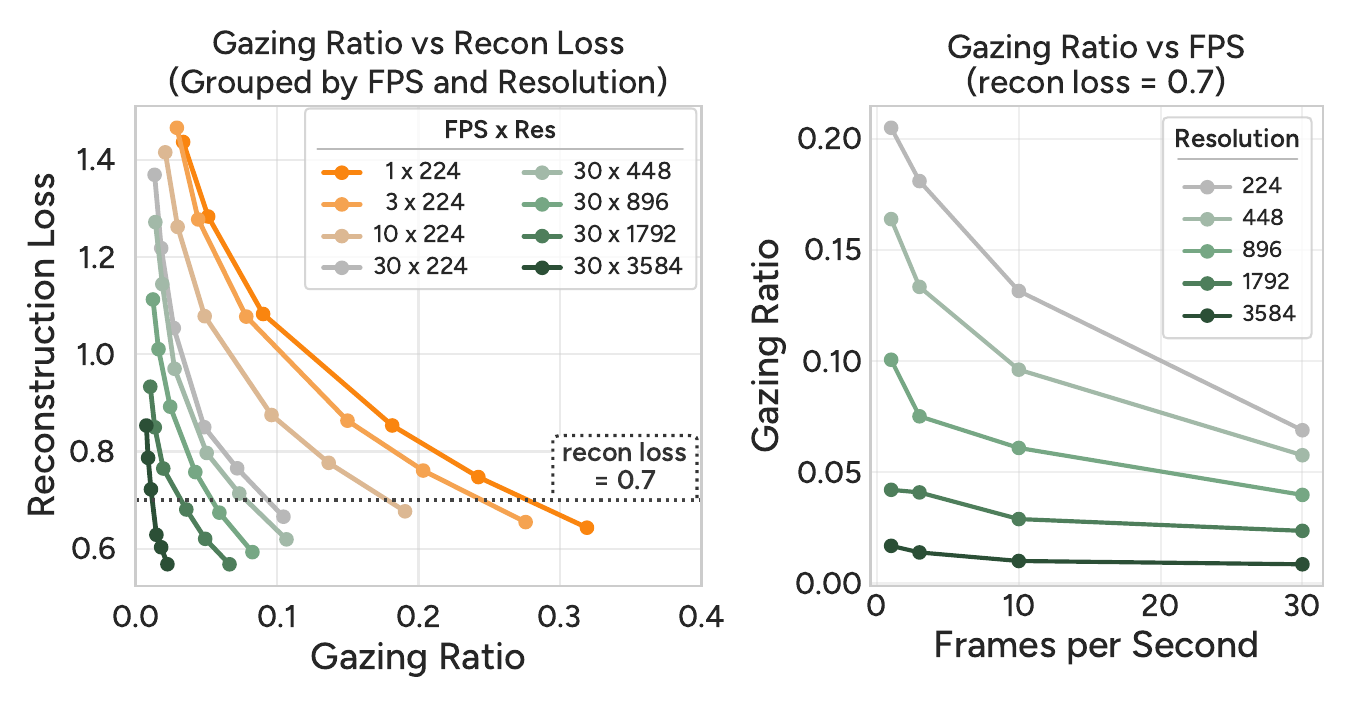}
  \end{center}

  \vspace{-1em}
  \caption{\textbf{What gazing ratio do we need for different video types?} \textit{(Left)} There is a trade-off between gazing ratio and reconstruction loss: videos with higher FPS or resolution need lower gazing ratio to reach the same reconstruction quality. \textit{(Right)} Gazing ratios required to reach a loss of 0.7 for videos with different FPS and resolution. 30-FPS, 4K-res videos only need $\sim$1\% patches.} 
  \label{fig:gazing_ratio_analysis}
\end{figure}

\model's efficiency comes from selecting only a small fraction of patches --- but does it make principled decisions about \textit{which} patches to select and at \textit{what} scale? We examine the factors that influence the behavior of \model and its generalization to videos with unseen styles and semantics.

\minisection{\model gazes more at moving patches.}
Motion is a primary source of new information across video frames, and thus should intuitively be selected (examples are shown in Fig. \ref{fig:gazing_visualization}). As illustrated in Fig. \ref{fig:gazing_flow}, \model does indeed prioritize motion: tested on pairs of videos and flow data from FlyingChairs~\citep{DFIB15,ISKB18}, we find that across all scales, it more frequently selects patches with higher optical flow.

\minisection{\model uses finer scales for more detailed patches.}
Regions with different detailedness should be represented with different scales, as illustrated in Fig. \ref{fig:gazing_visualization}. To verify this, we measure the relationship between gazing scale and patch detail by convolving 2,000 ImageNet images~\citep{5206848} with a Laplacian kernel and computing variance over each patch (higher values indicate more detail). Fig. \ref{fig:gazing_texture} (left) shows that at finer scales, \model tends to select more detailed patches. Fig. \ref{fig:gazing_texture} (right) confirms that \model gazes at higher resolutions to capture fine detail.

\minisection{\model generalizes to OOD videos.}
We test whether \model transfers beyond its training distribution to unseen semantics and styles, as shown in Fig. \ref{fig:ood}.
First, we show that \model behavior holds in unconventional scenarios including a CCTV footage, a robot video, and a video~\cite{radosavovic2024humanoid} that constantly swaps its foreground object between human, gorilla, and humanoid robot (created with Luma's Ray2 Flash). In each example, \model successfully tracks changing regions despite the novel semantics or unexpected changes.
Next, we test on unseen styles by style-transferring a video with TokenFlow~\citep{qu2025tokenflowunifiedimagetokenizer} to vary texture and global illumination. Across styles, \model maintains consistent gazing patterns, continuing to track the falling subject.

\begin{figure*}[t]
  \begin{center}
    \includegraphics[width=0.478\textwidth]{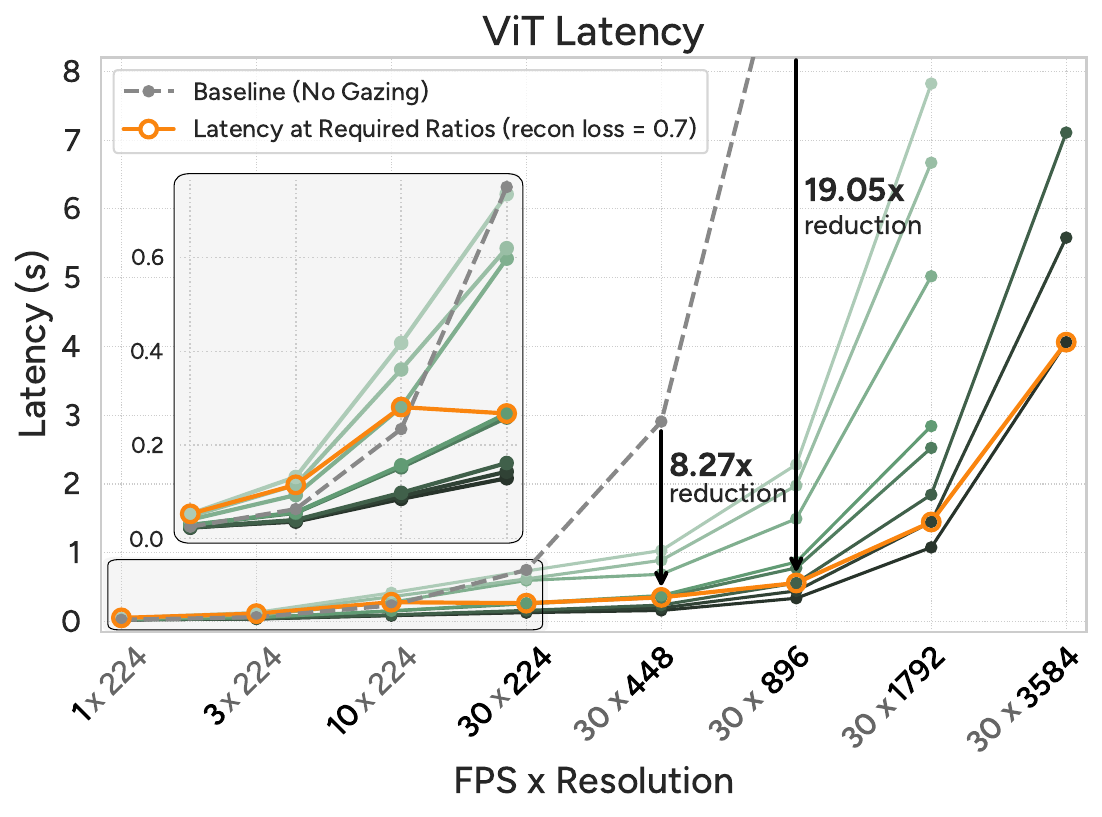}
    \includegraphics[width=0.517\textwidth]{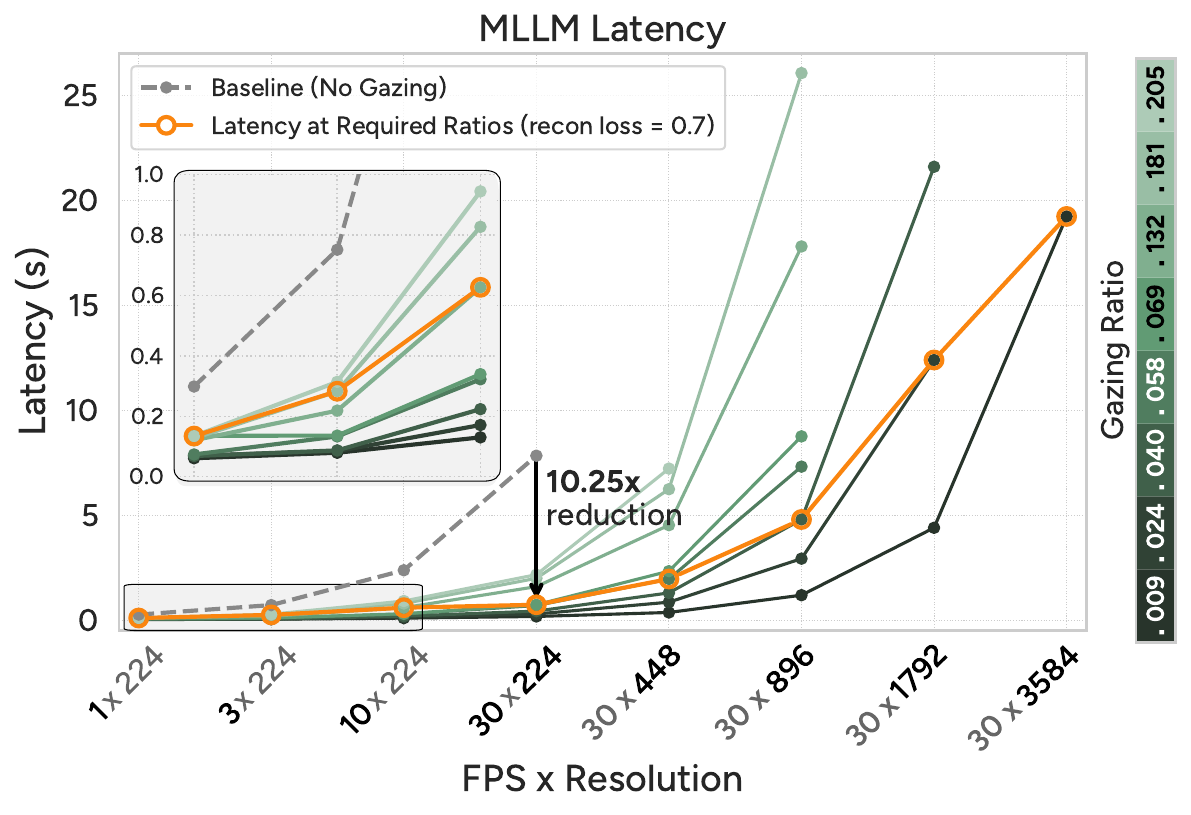}
  \end{center}
  \vspace{-0.5em}
  \caption{\textbf{Efficiency gain on ViTs and MLLMs with \model.} We benchmark the ViT and MLLM latency of encoding one second of video with varying FPS and resolution. \model can select different numbers of patches to vary latency depending on user needs. When using the gazing ratio required for a reconstruction loss of 0.7, \model reduces the ViT and MLLM latency by up to 19$\times$ and 10$\times$.} 
  \label{fig:efficiency_analysis}
\end{figure*}
\begin{figure*}[t]
  \begin{center}
    \includegraphics[width=1.0\textwidth]{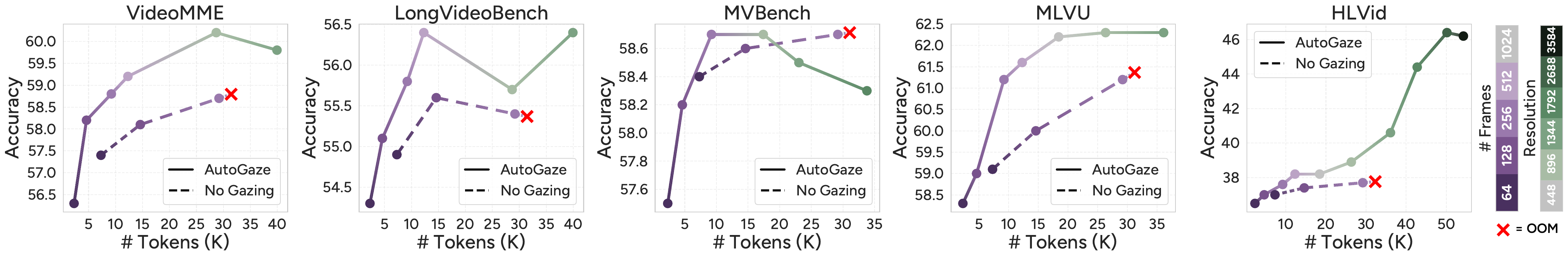}
  \end{center}
  \vspace{-0.5em}
  \caption{\textbf{Scaling MLLMs to more video tokens. }\model enables scaling to longer videos and higher resolution, while the baseline runs out of memory beyond 256 frames. Performance boosts are especially clear with \benchmark, where high-resolution video processing is required.
  } 
  \label{fig:scaling_property}
\end{figure*}

\subsection{Efficiency of ViT and MLLM with \model}
\label{sec:exp_efficiency}

We now study how efficient ViTs and MLLMs can be by selecting fewer patches via \model. To answer this question, we first analyze the number of patches required to represent a video with \model, and then benchmark the latency of ViT and MLLM when only the selected patches are processed.

\minisection{How many patches do we need to represent a video?} The number of patches needed depends on both the required reconstruction loss and the level of redundancy (e.g., different FPS and resolution) in the video. We first pinpoint the reconstruction loss that leads to minimal performance drop in downstream MLLMs, and find that a threshold of 0.7 usually leads to less than 0.5\% performance degradation across benchmarks (see detailed results in Appendix~\ref{appendix_sec:quantitative_result}). Next, we analyze how many patches are needed to represent videos with varying FPS and resolutions in order to achieve a reconstruction loss of 0.7. Fig. \ref{fig:gazing_ratio_analysis} (Left) shows the reconstruction loss for different gazing ratio and different FPS and resolution. We can see the gazing ratio required for a certain loss decreases with higher FPS and resolution. 
Fig. \ref{fig:gazing_ratio_analysis} (Right) shows complete results of gazing ratios required to reach a loss of 0.7 for different videos. Usually a video can be represented with 4$\times$-100$\times$ fewer patches. Specifically, only $\sim$1\% patches are needed for 30-FPS, 4K videos.

\minisection{How much faster are ViTs and MLLMs with \model?} With a target reconstruction loss of 0.7, we analyze the efficiency gains by testing wall-clock ViT and MLLM latency when processing one second of video. We use FP32 and disable flash attention for all models.
We report the aggregated latency of \model and ViT / MLLM, and compare to the baseline without gazing in Fig. \ref{fig:efficiency_analysis}. The ViT baseline quickly runs out of memory around 30 FPS and 896 resolution, and the MLLM baseline can only encode 30 FPS and 224 resolution. In contrast, \model helps efficiently process videos with lower gazing ratios. When using the gazing ratio required for a reconstruction loss of 0.7, it achieves up to 19$\times$ and 10$\times$ speedup for ViTs and MLLMs respectively, enabling scaling to 4K resolution.

\begin{figure}[h]
  \begin{minipage}[c]{0.55\columnwidth}
    \centering
  \includegraphics[width=\linewidth]{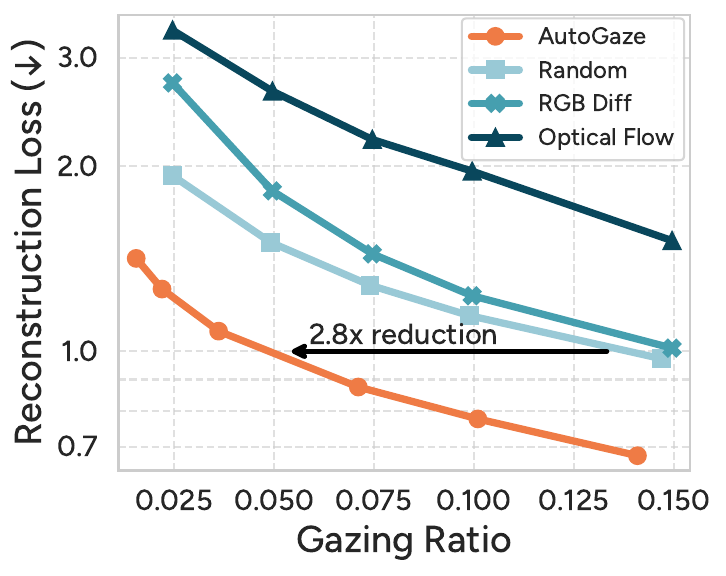}
  \end{minipage}\hfill
  \begin{minipage}[c]{0.42\columnwidth}
    \raisebox{5em}{%
    \begin{minipage}{\linewidth}
  \caption{\textbf{Comparison to baseline gazing methods.} \model can use smaller gazing ratios to reach the same reconstruction loss, compared to other heuristics-based gazing approaches. }
    \label{fig:compare_to_gaze_baseline}
  \end{minipage}
  }
  \end{minipage}
\end{figure}

\subsection{Scaling MLLMs with \model}
\label{sec:exp_performance}

\begin{table*}[t]
\caption{\textbf{Comparison to state-of-the-art MLLMs.} NVILA-8B-Video with \model is scaled to 1K-frame, 4K-resolution videos, achieving competitive performance on general and long video benchmarks and state-of-the-art result on \benchmark.
}
\label{tab:sota}
\centering

\resizebox{\linewidth}{!}{%
\begin{tabular}{lllcccccccc}
\toprule
 &  &  & \multicolumn{4}{c}{\textit{general video}} & \multicolumn{3}{c}{\textit{long video}} & \textit{high-res \& long} \\
\cmidrule(lr){4-7}\cmidrule(lr){8-10}\cmidrule(lr){11-11}
Models & \makecell[l]{Max \\ \#F} & \makecell[l]{Max \\ Res.} & \makecell{VideoMME \\ (w/o sub)} & \makecell{VideoMME \\ (w/ sub)} & \makecell{MVBench \\ (test)} & \makecell{NExT-QA \\ (mc)}  & \makecell{L-VidBench \\ (val)} & \makecell{EgoSchema \\ (test)} & \makecell{MLVU \\ (m-avg)} & \makecell{\benchmark \\ (test)} \\ 
\midrule
Gemini 1.5-Pro~\cite{team2024gemini} & - & - & 75.0 & 81.3 & 60.5 & - & 64.0 & 71.2 & - & - \\
Gemini 2.5 Flash-Lite~\cite{comanici2025gemini} & - & - & 65.0 & - & - & - & - & - & 69.3 & 52.2 \\
GPT-4o~\cite{hurst2024gpt} & - & - & 71.9 & 77.2 & 64.6 & - & 66.7 & 72.2 & 64.6 & 49.3 \\
\midrule
LLaVA-OV-8B~\cite{li2024llava} & 32 & 384 & 58.2 & 61.5 & 56.7 & 79.4 & 56.5 & 60.1 & 64.7 & 41.1 \\
LongVILA-7B~\cite{chen2024longvila} & 2048 & 384 & 60.1 & 65.1 & 67.1 & 80.7 & 57.1 & - & - & 41.4 \\
LongVILA-R1-7B~\cite{chen2025scaling} & 8192 & 448 & 65.1 & 71.1 & - & 81.5 & 58.0 & - & - & 42.2 \\
Apollo-7B~\cite{zohar2025apollo} & 2FPS & 384 & 61.3 & 63.3 & - & - & 58.5 & - & 70.9 & - \\
VideoLLaMA3-7B~\cite{zhang2025videollama} & 180 & 384 & 66.2 & 70.3 & 69.7 & \textbf{84.5} & 59.8 & 63.3 & 73.0 & 38.8 \\
VideoChat-Flash~\cite{li2024videochat} & 10000 & 448 & 65.3 & 69.7 & \textbf{74.0} & - & \textbf{64.7} & - & \textbf{74.7} & 46.6 \\
InternVL3.5-8B~\cite{wang2025internvl3} & 64 & 448 & 66.0 & 68.6 & 72.1 & - & 62.1 & - & 70.2 & 39.9 \\
Qwen2.5-VL-7B~\cite{bai2025qwen2} & 48 & 896 & 65.1 & 71.6 & 69.6 & - & 56.0 & 65.0 & 70.2 & 48.1 \\
\midrule
NVILA-8B-Video~\cite{liu2025nvila} & 256 & 448 & 64.2 & 70.0 & 68.1 & 82.2 & 57.7 & - & 70.1 & 42.5 \\
\rowcolor{Gray!20} \ + \model & 1024 & 3584 & \textbf{67.0} & \textbf{71.8} & 69.7 & 82.8 & 61.0 & \textbf{66.9} & 71.6 & \textbf{52.6} \\
\rowcolor{Gray!20} \textit{(vs. NVILA-8B-Video)} & \textcolor{darkgreen!70}{\textbf{\textit{($\times$4)}}} & \textcolor{darkgreen!70}{\textbf{\textit{($\times$8)}}} & \textcolor{darkgreen!70}{\textbf{\textit{(+2.8)}}} & \textcolor{darkgreen!70}{\textbf{\textit{(+1.8)}}} & \textcolor{darkgreen!70}{\textbf{\textit{(+1.6)}}} & \textcolor{darkgreen!70}{\textbf{\textit{(+0.6)}}} & \textcolor{darkgreen!70}{\textbf{\textit{(+3.3)}}} & \textcolor{darkgreen!70}{\textbf{\textit{-}}} & \textcolor{darkgreen!70}{\textbf{\textit{(+1.5)}}} & \textcolor{darkgreen!70}{\textbf{\textit{(+10.1)}}} \\
\bottomrule
\end{tabular}
}
\end{table*}

Leveraging \model's efficiency, we scale MLLMs to longer, higher-resolutions videos and achieve state-of-the-art performance on video benchmarks.

\minisection{Scaling properties.} We compare performance and efficiency when scaling MLLMs at test time to longer and high-resolution videos with or without \model, and report results in Fig. \ref{fig:scaling_property}. We first scale the number of frames, identify the best frame count for each benchmark, then scale resolution. Starting from 64 frames and 448 resolution, MLLM with \model has slightly worse performance than the baseline while using $\sim$4$\times$ fewer tokens. This performance drop vanishes after scaling to 256 frames. When further scaling video duration and resolution, the baseline runs out of memory while \model enables scaling to 1K frames and 4K resolution with consistent improvements. 
Note that on some benchmarks, using too long or too high-resolution videos is detrimental, likely because those benchmarks require neither, while scaling to 4K resolution significantly improves performance on \benchmark, verifying it does require high resolution.

\minisection{Comparing to state-of-the-art MLLMs.} We train NVILA-8B-Video~\cite{liu2025nvila} with \model at 256 frames and 896 resolution, scale up to 1K frames and 4K resolution at test time, and compare to existing MLLMs in Tab. \ref{tab:sota}.
Using \model, it consistently improves over the base NVILA-8B-Video on all benchmarks. Notably, performance on \benchmark improves by 10.1\%, surpassing all open-source MLLMs like Qwen2.5-VL-7B~\cite{bai2025qwen2} and proprietary models like GPT-4o~\cite{hurst2024gpt} despite much smaller model size or training dataset. NVILA-8B-Video with \model also outperforms others on VideoMME, though falls short on MVBench and LongVideoBench likely due to differing training recipes.

\subsection{Comparing to Token Pruning Baselines}
\label{sec:exp_baseline_comparison}

\begin{table}[t]
\vspace{1em}
\caption{\textbf{Comparing \model with MLLM token reduction methods.} We compare to baselines of spatial/temporal/spatiotemporal (S-/T-/ST-) token reduction approaches that are either prompt-agnostic (PA) or prompt-dependent (PD). All the methods select 6.25\% visual tokens on average.}
\label{tab:compare_to_token_reduction_baseline}
\centering
\begin{footnotesize}
\begin{tabular}{llcccc}
\toprule
Type & Method & \makecell{ViT \\ lat.} & \makecell{LLM \\ lat.} & \makecell{V-MME \\ (w/o sub)} & \makecell{L-Vid \\ (val)} \\ 
\midrule
- & No Reduction & 2.20s & 1.42s & 53.4 & 51.1 \\
\midrule
\multirow{3}{*}{S-PA} & S-Pool & 2.20s & 0.18s & 51.5 & 47.2 \\
 & ToMe~\cite{bolya2022token} & 2.23s & 0.11s & 51.5 & 49.3 \\
 & VisionZip~\cite{yang2025visionzip} & 2.22s & 0.15s & 50.7 & 48.5 \\
\midrule
S-PD & FastV~\cite{chen2024image} & 2.23s & 0.38s & \textbf{53.0} & 46.3 \\
\midrule
T-PA & T-Pool & 2.20s & 0.13s & 52.2 & 50.0 \\
\midrule
T-PD & AKS~\cite{tang2025adaptive} & 3.27s & 0.12s & 50.8 & 49.5 \\
\midrule
\multirow{4}{*}{ST-PA} & ST-Pool & 2.19s & 0.13s & 52.0 & 49.8 \\
 & STORM~\cite{jiang2025storm} & 2.18s & 0.15s & 52.7 & \textbf{51.5} \\
 & FastVID~\cite{shen2025fastvid} & 2.34s & 0.12s & 52.4 & 50.3 \\
 & F-16~\cite{li2025improving} & 2.20s & 0.18s & 51.8 & 50.0 \\
\midrule
\multirow{3}{*}{ST-PD} & LongVU~\cite{shen2024longvu} & 2.17s & 0.12s & 52.2 & 50.1 \\
 & PruneVID~\cite{huang2024prunevid} & 2.52s & 0.15s & 50.3 & 48.0 \\
 & VChat-Flash~\cite{li2024videochat} & 2.21s & 0.15s & 52.4 & 49.9 \\
\midrule
\rowcolor{Gray!20} ST-PA & \model & \textbf{0.55s} & \textbf{0.10s} & 52.3 & 50.3 \\
\bottomrule
\end{tabular}
\end{footnotesize}
\end{table}

\minisection{Comparing to baseline gazing approaches.} 
In Fig. \ref{fig:compare_to_gaze_baseline}, we compare \model to heuristic-based gazing approaches that select patches randomly (\textbf{Random Gaze}), with the largest RGB difference (\textbf{RGB-Diff Gaze}) or with the largest optical flow using SEA-RAFT~\cite{wang2024sea} (\textbf{Optical-Flow Gaze}).
We left-pad a blank image when computing differences for the first frame. 
\model greatly improves efficiency --- for example, reaching reconstruction loss 1.0 with 5\% patches versus 15\% for Random Gaze.
We find RGB-Diff Gaze and Optical-Flow Gaze to be worse than Random Gaze because they fixate on the first frame (an abrupt change from the padding) and ignore other frames, causing information loss.

\minisection{Comparing to MLLM token reduction baselines.} We compare \model to existing MLLM token reduction methods~\cite{bolya2022token,yang2025visionzip,chen2024image,tang2025adaptive,jiang2025storm,li2025improving,shen2025fastvid,shen2024longvu,huang2024prunevid,li2024videochat} and three simple baselines: spatial, temporal, and spatiotemporal pooling (S-Pool, T-Pool, and ST-Pool). These approaches process the whole video in the ViT, then reduce visual tokens in the LLM. We use 128-frame videos and 6.25\% selection ratio and report results in Tab. \ref{tab:compare_to_token_reduction_baseline}. The baseline without token reduction has high ViT and LLM latency, with ViT latency slightly higher than LLM latency despite its smaller size due to LLMs' token shuffling~\cite{liu2025nvila,bai2025qwen2}. While baseline methods improve LLM latency by 3.7$\times$-13.4$\times$, the ViT latency remains unchanged. In contrast, \model significantly improves ViT latency by 4$\times$ in addition to the LLM latency improvement. Beyond speedups, both the token-reduction baselines and \model retain performance comparable to the no-reduction baseline.

\subsection{Ablations}
\label{sec:exp_ablation}

\begin{table}[t]
\vspace{1em}
\caption{\textbf{Ablation on \model training pipeline.} Both NTP pre-training and RL post-training helps with the performance.}
\label{tab:ablation_training_pipeline}
\centering
\begin{small}
\begin{tabular}{cccc}
\toprule
Pre-Train & Post-Train & Recon Loss & Gazing Ratio \\ 
\midrule
\xmark & \xmark & 0.7 & 0.263 \\
\cmark & \xmark & 0.7 & 0.102 \\
\xmark & \cmark & 0.7 & 0.209 \\
\rowcolor{Gray!20} \cmark & \cmark & 0.7 & \textbf{0.094} \\
\bottomrule
\end{tabular}
\end{small}
\end{table}

\begin{table}[t]
\caption{\textbf{Ablations of \model model designs.}}
\label{tab:ablation_model_design}
\centering
\begin{small}
\begin{tabular}{ccccc}
\toprule
\cellcolor{Gray!5} \cellcolor{Gray!20} \makecell{Multi-Token \\ Pred.} & \cellcolor{Gray!40} \makecell{Multi-Scale \\ Gazing} & \makecell{Recon \\ Loss} & \makecell{Gazing \\ Ratio} & Latency \\ 
\midrule
\cellcolor{Gray!20} 1 & \cmark & 0.7 & 0.074 & 0.949s \\
\cellcolor{Gray!20} 5 & \cmark & 0.7 & 0.078 & 0.246s \\
\cellcolor{Gray!20} 10 & \cmark & 0.7 & \textbf{0.094} & \textbf{0.193s} \\
\cellcolor{Gray!20} 20 & \cmark & 0.7 & 0.109 & 0.156s \\
\midrule
10 & \cellcolor{Gray!40} \cmark & 0.7 & \textbf{0.094} & \textbf{0.193s} \\
10 & \cellcolor{Gray!40} \xmark & 0.7 & 0.220 & 0.467s\\
\bottomrule
\end{tabular}
\end{small}
\end{table}

\minisection{Training pipeline of \model.} We analyze the effect of the two-stage training pipeline by comparing the gazing ratio required to reach reconstruction loss 0.7 when removing different training stages. As shown in Tab. \ref{tab:ablation_training_pipeline}, with only pre-training or post-training, efficiency improves over no training, while pre-training contributing more. With both stages, we obtain the lowest gazing ratio, achieving $\sim$10\% improvement over pre-training only.

\minisection{Model designs of \model.} We analyze the effect of multi-token prediction and multi-scale gazing on the model performance and efficiency (Tab. \ref{tab:ablation_model_design}). We report the gazing ratio required to reach reconstruction loss 0.7 and the corresponding latency when processing a 1-second video at 10 FPS and 224 resolution. 
We find decoding more tokens at a time leads to lower latency but higher gazing ratio, with decoding 10 tokens balancing both factors well. On the other hand, multi-scale gazing reduces the gazing ratio and improves the efficiency by 2.3$\times$.

%% file: sections/5_discussion.tex
\section{Conclusion}

We introduce \model, a lightweight framework that removes redundant video patches to improve ViT and MLLM efficiency. Trained via NTP on gazing sequences collected through a greedy algorithm and RL with reconstruction reward, \model learns to select a minimal set of multi-scale patches that reconstructs the video within a user-specified threshold. Empirically, \model reduces visual tokens by 4$\times$–100$\times$ and accelerates ViTs and MLLMs by up to 19$\times$ and 10$\times$, enabling 1024-frame 4K-resolution video understanding and improving performance on video benchmarks. We further introduce \benchmark, the first long-form (5-minute), high-resolution (4K) video QA benchmark, where an MLLM with \model outperforms previous SOTA model by 4.5\%.

%% file: sections/appendix.tex
\section{Additional Details of \model Model Design}
\label{appendix_sec:model_detail}

\minisection{Model architecture.} \model contains a convolutional vision encoder, a visual connector, and a transformer decoder. The convolutional encoder contains one 2D convolutional layer with spatial kernel size of 16 to embed each patch, and one 3D convolutional layer with spatial and temporal kernel sizes of 3 to extract the spatiotemporal vision features of each frame based on the current frame and previous two frames. Note that the vision encoder is causal. The visual connector bridges between the vision encoder output and the transformer decoder input, adding positional embeddings to the output visual features from the vision encoder and passing them to the decoder. The positional embeddings are added in each frame separately, such that each token in each frame is aware of its spatial position in the frame. The transformer decoder uses the same architecture design as LLaMA 3~\cite{dubey2024llama} but with only four layers. The decoder takes in each frame's visual tokens and decodes the patch indices. It also predicts the reconstruction loss of the current frame at each step using a linear decoding head. The vocabulary of the decoder only contains all the possible patch indices in a frame. We use four scales, i.e., 32$\times$32, 64$\times$64, 112$\times$112, and 224$\times$224. Since the patch size is 16, the number of all possible patches (i.e., the vocabulary size of the decoder) is 4 + 16 + 49 + 196 = 265. The hidden dimension of the whole model is 192.

\minisection{Instantiation of the reconstruction objective.} \model is trained to select as few patches as possible while keeping a certain level of reconstruction loss (Eq. \ref{eq:objective}). The reconstruction objective includes a distance function $L(\cdot, \cdot)$ and a video reconstruction model $\texttt{Recon}(\cdot)$. We obtain the reconstruction model by taking an MAE~\citep{he2022masked} pre-trained on images and fine-tuning it on videos with the same masked auto-encoding objective such that it can reconstruct a video from partially observed patches. The resulting model is similar to VideoMAE~\citep{tong2022videomae} except that the self-attention layers are block-causal, \ie, the model reconstructs each frame based on only the current and previous frames. This is important because the gazing model is also causal and we should not train it to optimize a reconstruction loss that depends on the future. This also allows us to calculate the reconstruction loss of each frame based on previous gazed patches up until any step at that frame, which we use to supervise the reconstruction loss prediction at each step. For the distance function, we use a combination of pixel reconstruction loss and perceptual loss~\citep{zhang2018unreasonable,johnson2016perceptual}, \ie, a weighted sum of $\ell_1$ loss in the pixel space and $\ell_2$ loss on the frame-wise DINOv2~\citep{oquab2023dinov2} and SigLIP2~\citep{tschannen2025siglip} embeddings between the reconstructed and the original video. The weights for $\ell_1$ loss, DINOv2 embedding loss, and SigLIP2 embedding loss are 1, 0.3, and 0.3 respectively.

\section{Additional Details of \model Training Pipeline}
\label{appendix_sec:training_detail}

\minisection{NTP pre-training.} We pre-train \model on about 250K videos with paired gazing sequences. We train for 150 epochs, with a batch size of 256 and a learning rate of 5e-4.

\minisection{RL post-training.} We use GRPO algorithm that is a simplified and on-policy version of the original version, i.e., we remove the KL regularization term and use the same policy at each step as the old policy in the original algorithm. We use a group size of 12. When calculating the advantage for each gazing step, we count in the negative reconstruction loss for the current frame as well as the subsequent frames. The reconstruction loss for each frame is discounted based on the number of gazing steps between the current gazing and the last gazing of that frame. The discounting factor we use is 0.995. We also follow Dr. GRPO to remove the std normalization and the sequence length normalizer in GRPO. During RL training, we anneal the temperature when rolling out the gazing prediction from 1 to 0.01. At each step of training, instead of running VideoMAE on all the frame to get the reconstruction reward for every frame, we instead just randomly sample 2 frames to reconstruct and only use the reconstruction rewards for those two frames for RL training in order to improve training efficiency. When rolling out gazing prediction during training, we set a reconstruction loss threshold of 0.7, let \model to predict one gazing sequence based on the threshold, record the number of gazing for each frame, and then roll out a group of gazing sequences with the same number of gazing for each frame for GRPO loss. This ensures every gazing sequence in the same GRPO group has the same number of gazing for each frame, such that the reconstruction rewards between sequences can be fairly compared. We train for 3 epochs, with a batch size of 256 and a learning rate of 5e-4.

\minisection{Training data.} For the training data, we first collect videos from datasets including Ego4D~\cite{grauman2022ego4d}, 100DoH~\cite{shan2020understanding}, and InternVid~\cite{wang2023internvid}, covering both exocentric and egocentric natural videos. We also create artificial videos that simulates camera motions by placing a window on an large image, slide the window to random directions, and take the content within the sliding window as a video. We create videos like this from high-resolution image datasets including SA-1B~\cite{kirillov2023segment} and IDL~\cite{biten2022ocr}, covering both natural and text-rich videos. In total, we end up with collecting $\sim$800K videos. For pre-training, we collect gazing sequences for a subset of $\sim$250K videos. Specifically, for each video, we first randomly sample the gazing ratios for each frame, and then search the best gazing sequence that optimizes the reconstruction loss under the sampled gazing ratio, using the greedy search algorithm described in Sec.~\ref{sec:autogaze_training}. When sampling the gazing ratio, we first randomly sample the average gazing ratio of the whole video from an exponential distribution between 0.02 and 0.2. Then, given the average gazing ratio of all the frames, we sample the gazing ratio for each frame using a Dirichlet distribution with alpha of 10 for the first frame and 3 for the rest of the frames, such that the gazing ratio is biased towards the first frame. This is to simulate the real distribution of gazing where the first frame is usually gazed at more since it contains completely new information without any history context.

\section{Additional Details of \benchmark Benchmark}
\label{appendix_sec:benchmark_detail}

\begin{figure*}
  \begin{center}
    \includegraphics[width=1.0\textwidth]{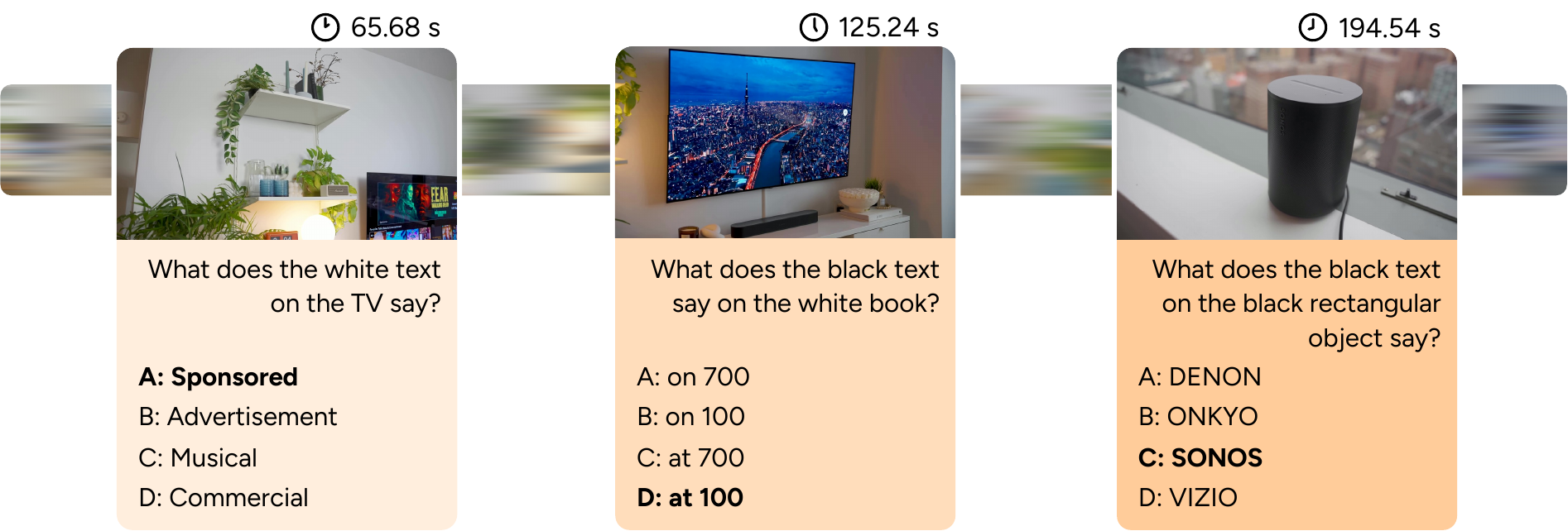} \\
    \vspace{3em}
    \includegraphics[width=1.0\textwidth]{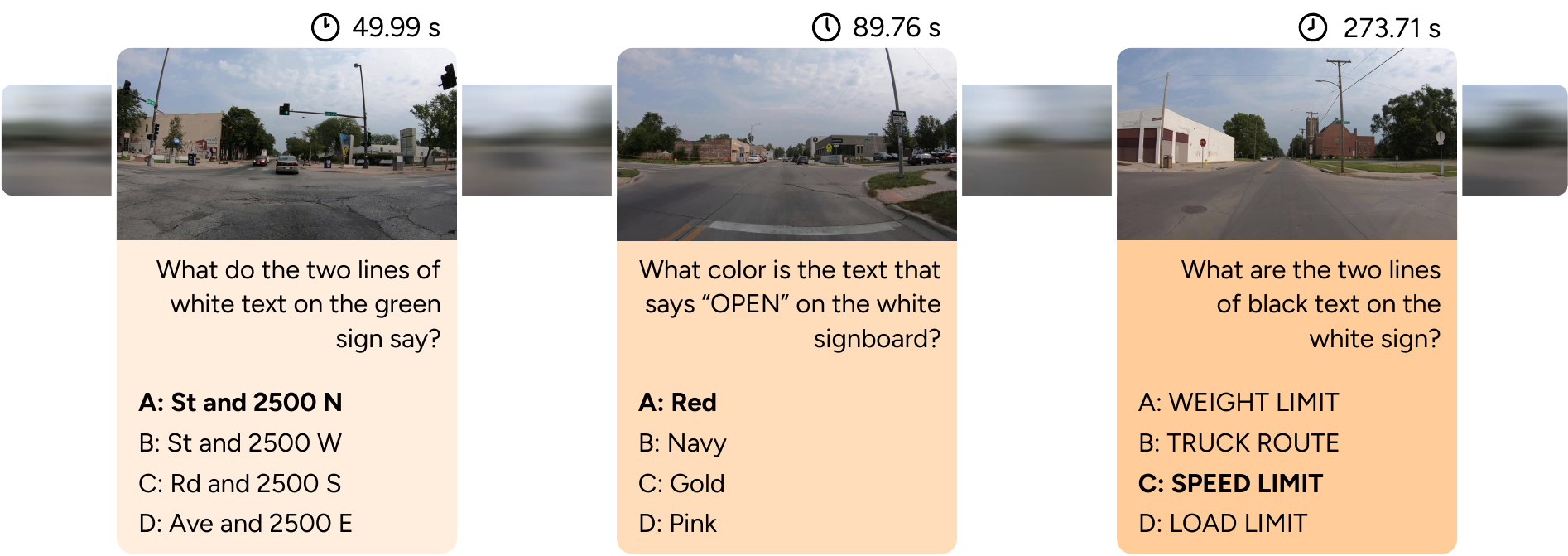}
  \end{center}
  \caption{\textbf{Examples of HLVid benchmark.} Each sample from HLVid consists of a 5-minute long video at 4K resolution, along with multiple-choice questions that require high-resolution video encoding to answer. The answers to HLVid's questions can be found at any spatiotemporal location, emphasizing long-context video understanding. Our video content is diverse and includes house tours (top), city driving (bottom), nature videos, etc.
  }
  \label{fig:hlvid_examples}
\end{figure*}

We collect 268 QA on videos of autonomous driving and household scenarios scraped from YouTube. Each video has 4K resolution and a large portion of them have 5 minutes of duration. We design the QAs such that every question needs high-resolution perception at at least 1K - 2K resolution to solve. We also review the QAs such that questions from the same video are asking about different details, and the answer to each question is not ambiguous, i.e., there is only one correct answer and there is no other correct answers appearing in other frames in the video. As as comparison, existing benchmarks mostly only focus of long videos but not high resolution. For example, LongVideoBench~\cite{wu2024longvideobench} and EgoSchema~\cite{mangalam2023egoschema} contain videos with an average duration of 473s and 180s separately, but their questions mostly only need low resolution to solve. Our benchmark, instead, requires both long and high-resolution video understanding to solve. See Figure \ref{fig:hlvid_examples} for examples of HLVid's videos and questions.

\section{Additional Details of Analyses}
\label{appendix_sec:analysis_detail}
Section \ref{sec:exp_analysis} explores what AutoGaze pays attention to in videos. Below, we provide further implementation details for the optical flow and patch detail analyses.

\minisection{Optical flow. } To analyze whether AutoGaze selects moving patches more often than static patches, we use the image pairs and optical flow map in the Flying Chairs dataset \cite{DFIB15}. Since the data only contains forward optical flow, it only assigns object motion to the pixels of the first frame, not the second frame. Therefore, we compute backward flow by finding each source pixel's resulting location and assigning the inverse optical flow there as well (occluded/invalid values are set to zero). The final flow map that we use is a pixel-wise maximum over the forward and backward optical flows. Given this map, we crop all possible patches at each gazing scale and compute the maximum optical flow in that patch.

\minisection{Patch detail. } We compute patch detail for Section \ref{sec:exp_analysis} using the variance of the Laplacian for each image. Specifically, we compute the patch ``detailedness'' by convolving each video frame with a 3x3 Laplace filter
$$
\begin{bmatrix}
0 & 1 & 0 \\
1 & -4 & 1 \\
0 & 1 & 0
\end{bmatrix}
$$ and computing the variance of the values in each possible patch. Lower variance corresponds to smoother/more constant coloring, while high variance corresponds to busier textures (\eg, stripes, text).

\section{Additional Quantitative Results}
\label{appendix_sec:quantitative_result}

\begin{figure}[h]
    \centering
    \includegraphics[width=1.0\linewidth]{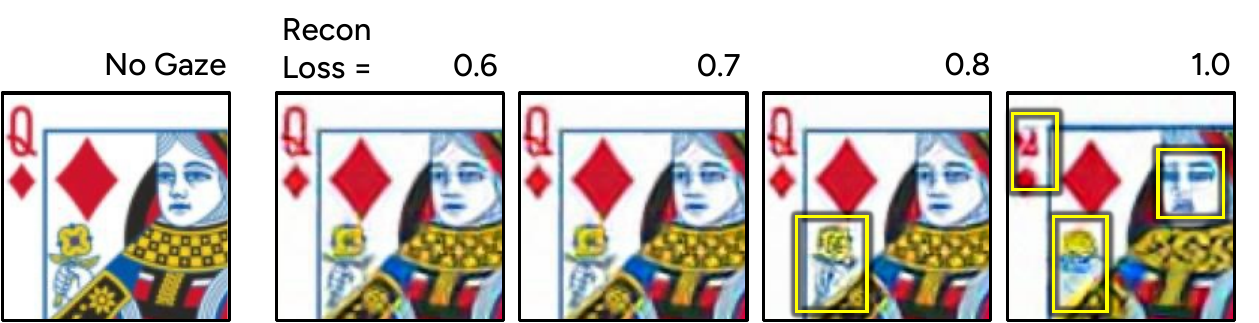}
    \caption{\textbf{Reconstruction quality under varying loss thresholds.} Outlined in \texttt{recon\_loss}$=0.8,1.0$ are noticeable visual defects.}
    \label{fig:recon_examples}
\end{figure}

\minisection{How large of a reconstruction loss is tolerable for downstream video understanding?} In Sec. \ref{sec:exp_efficiency}, we benchmark the efficiency of ViTs and MLLMs with \model. However, since the efficiency depends on how many patches to select for each video, which further depends on the reconstruction threshold, we first need to pinpoint the appropriate reconstruction loss threshold, i.e., the reconstruction loss threshold that leads to little or no downstream performance drop. To do this, we set different reconstruction loss thresholds and compare the MLLM performance. Tab. \ref{tab:performance_vs_recon_loss} shows results averaged over four benchmarks~\cite{fu2025video,wu2024longvideobench,li2024mvbench,xiao2021next}. Compared to the baseline without gazing, reconstruction loss below 0.7 leads to performance drop consistently less than 0.5\% across videos with different numbers of frames. This is supported by reconstruction visualization under different losses (Fig. \ref{fig:recon_examples}), where a loss $>0.7$ usually results in visible artifacts. Balancing performance and efficiency, we choose 0.7 as the threshold for MLLM experiments. 

\begin{table}[h]
\caption{\textbf{What reconstruction loss is tolerable for MLLM performance?} We find a loss of 0.7 has little performance drop.}
\label{tab:performance_vs_recon_loss}
\centering
\begin{small}
\begin{tabular}{cccc}
\toprule
Recon. Loss & 64 Frames & 128 Frames & 256 Frames \\ 
\midrule
No Gaze & 59.1 & 60.1 & 60.5  \\
\midrule
0.6 & 59.0 & 60.0 & 60.7\\
\rowcolor{Gray!20} 0.7 & 58.6 & 59.7 & 60.3 \\
0.8 & 57.8 & 58.4 & 59.0 \\
1.0 & 56.3 & 56.7 & 57.2 \\
\bottomrule
\end{tabular}
\end{small}
\end{table}

\begin{figure*}[t]
  \begin{center}
    \includegraphics[width=1.0\textwidth]{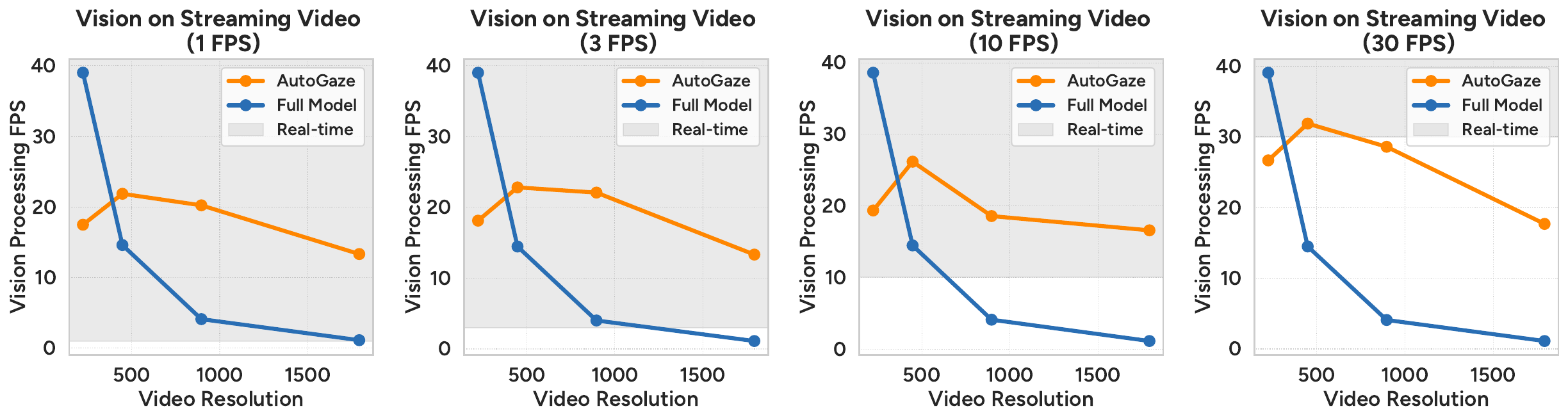}
    \includegraphics[width=1.0\textwidth]{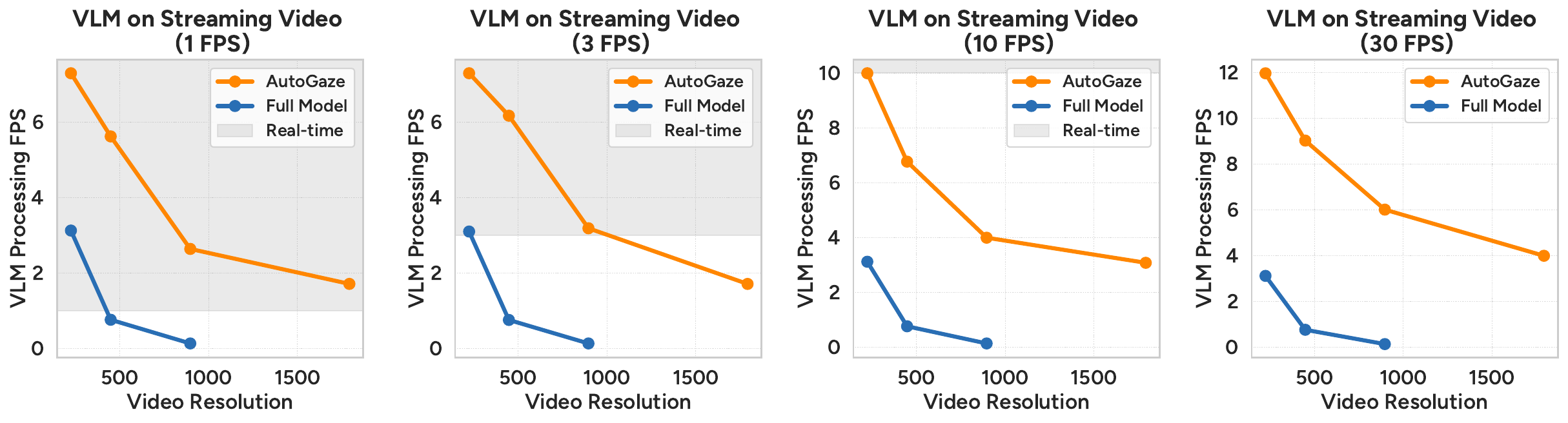}
  \end{center}
  \caption{\textbf{Efficiency on streaming videos.} We compare the ViT and MLLM efficiency on streaming videos with or without \model. We measure the maximum FPS of processing different types of videos (i.e., different FPS and resolution). For each plot, the gray area denotes the region where real-time processing of the video is achieved.
  }
  \label{fig:efficiency_analysis_streaming_video}
\end{figure*}

\minisection{Efficiency gain of ViTs and MLLMs on streaming videos.} In Sec. \ref{sec:exp_efficiency}, we benchmark the ViT and MLLM efficiency with \model. However, the benchmarking is conducted on static videos, i.e., the models can see the full video beforehand such that it can split each video into multiple clips and process in a batchified way. However, lots of real-world applications calls for streaming video understanding, i.e., the ViT and MLLM needs to process frames one by one. To this end, we also benchmark the ViT and MLLM efficiency under this situation, where they process frames one by one and we measure the maximum FPS they can process frames at. Results are shown in Fig. \ref{fig:efficiency_analysis_streaming_video}. We test on videos with different FPS and resolution, and compare the maximum FPS of ViT/MLLM with or without \model. We can see that ViTs and MLLMs with \model usually achieve an FPS that is up to $\sim$16$\times$ higher. This enables real-time processing for ViTs and MLLMs which is infeasible without \model (e.g., real-time ViT encoding of 10FPS videos at resolution higher than 500, and real-time MLLM processing of 3 FPS videos at 1K resolution).

\section{Additional Qualitative Results}
\label{appendix_sec:qualitative_result}

\begin{figure*}
    \centering
    \includegraphics[width=0.94\linewidth]{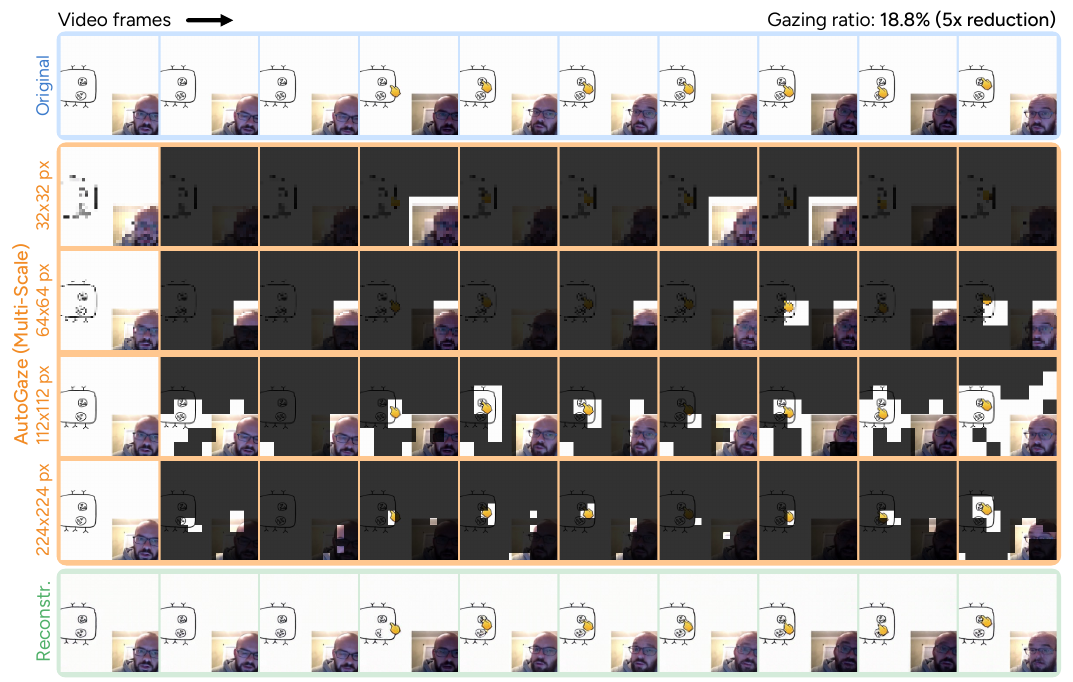}
    \caption{\textbf{Picture-in-picture whiteboard lecture.} After the first frame, \model focuses on the moving cursor and the lecturer's face.}
    \label{fig:recon_examples_app_1}
\end{figure*}

\begin{figure*}
    \centering
    \includegraphics[width=0.94\linewidth]{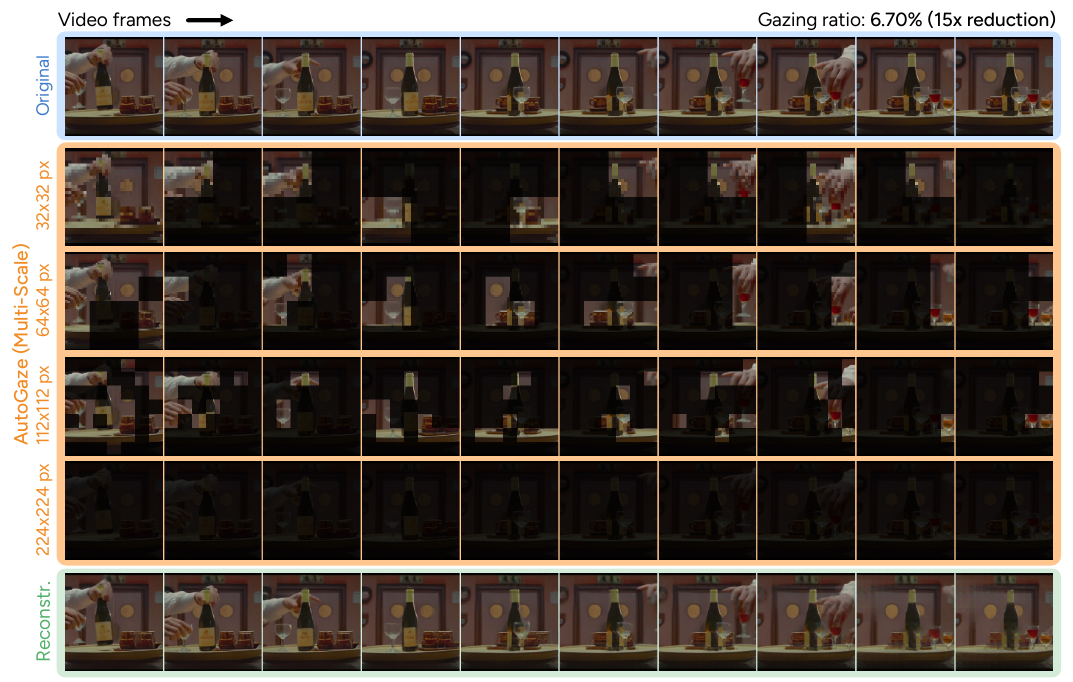}
    \caption{\textbf{Film clip.} \model selects minimal patches to track the drink bottles and glasses as they are moved by hands a rotating plate.}
    \label{fig:recon_examples_app_2}
\end{figure*}

\begin{figure*}
    \centering
    \includegraphics[width=0.94\linewidth]{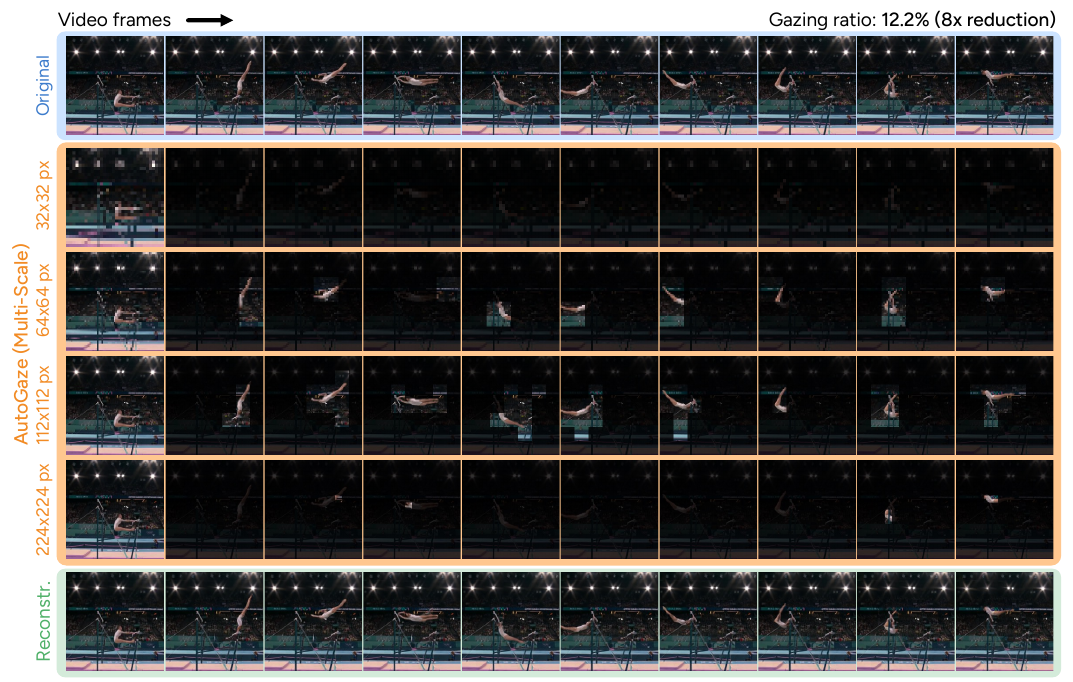}
    \caption{\textbf{Gymnastics clip.} \model tracks the gymnast across the uneven bars, using finer scales when appropriate.}
    \label{fig:recon_examples_app_3}
\end{figure*}

\begin{figure*}
    \centering
    \includegraphics[width=0.94\linewidth]{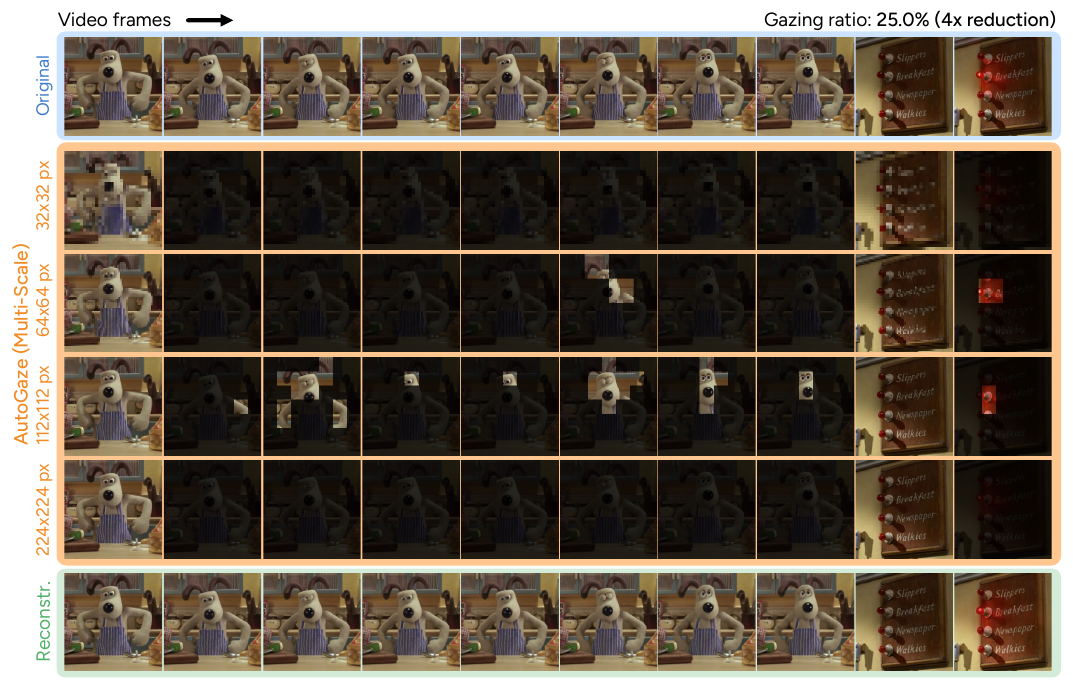}
    \caption{\textbf{Claymation cartoon.} \model captures small movements (blinking), scene changes, and enough patches to reconstruct text.}
    \label{fig:recon_examples_app_4}
\end{figure*}

\begin{figure*}
    \centering
    \includegraphics[width=0.94\linewidth]{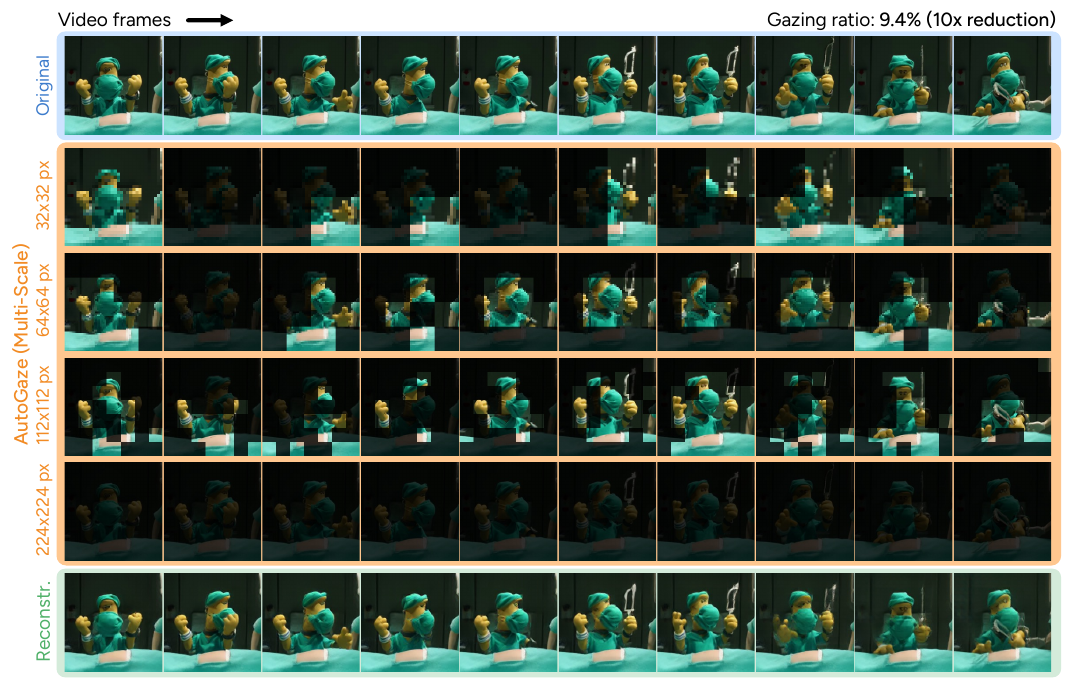}
    \caption{\textbf{Claymation cartoon.} \model skips the finest scale as it is not needed to achieve the specified reconstruction loss.}
    \label{fig:recon_examples_app_5}
\end{figure*}

\begin{figure*}
    \centering
    \includegraphics[width=0.94\linewidth]{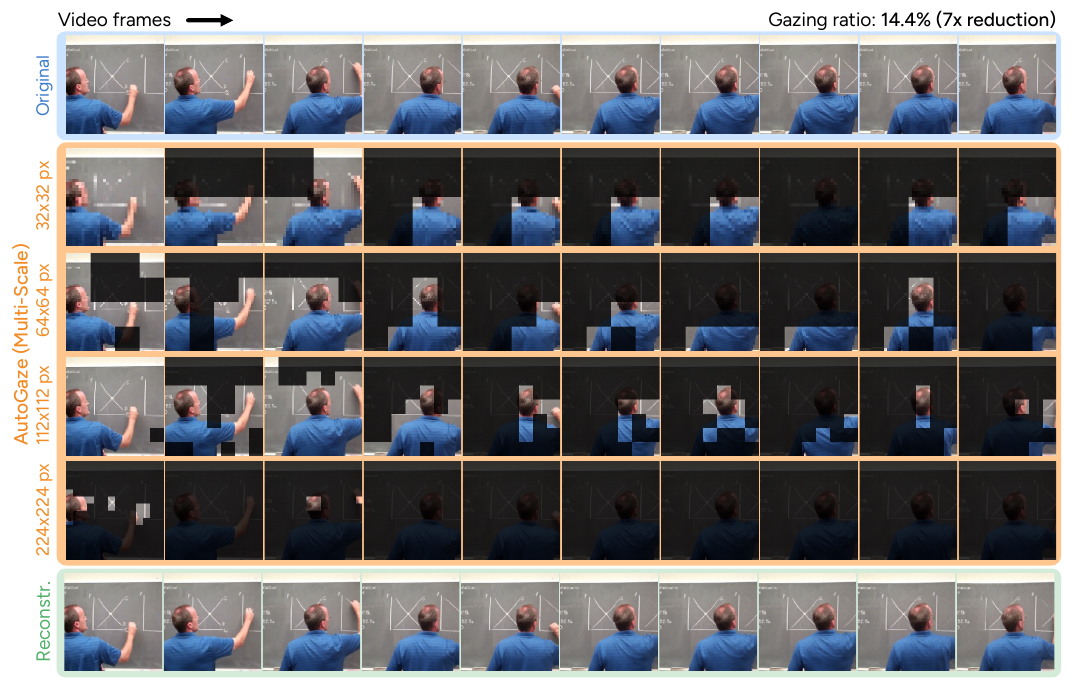}
    \caption{\textbf{Chalkboard lecture.} The minimal patches are selected to reconstruct both the lecturer's movement and the chalkboard writing.}
    \label{fig:recon_examples_app_6}
\end{figure*}

\begin{figure*}
    \centering
    \includegraphics[width=0.94\linewidth]{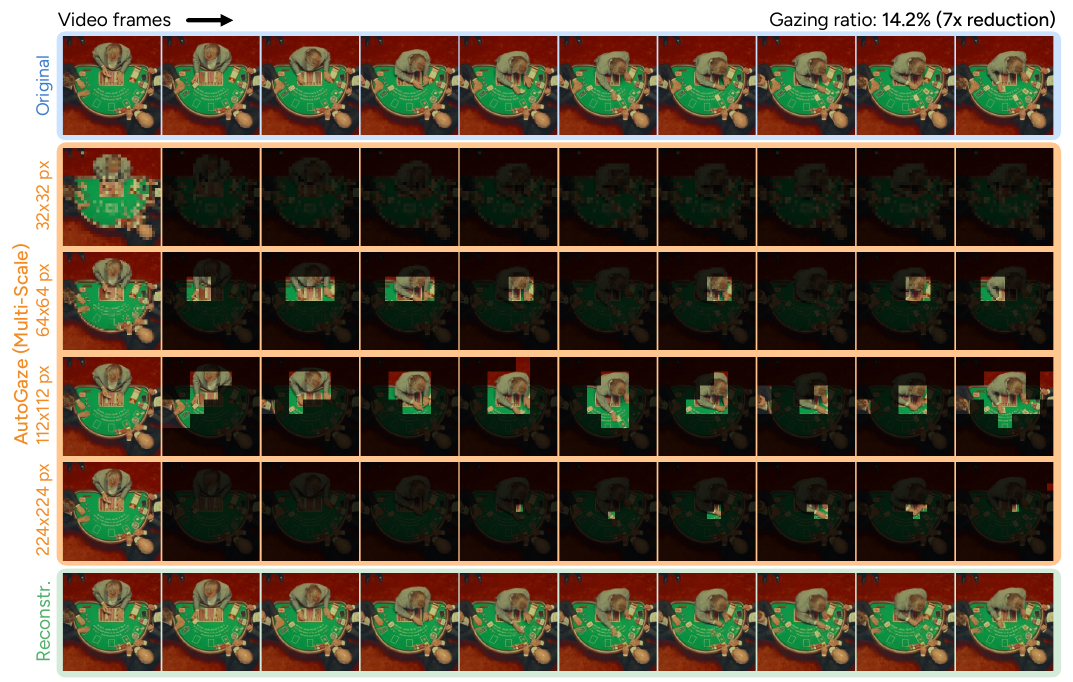}
    \caption{\textbf{Film clip.} In this game of Blackjack, \model uses finer scales to capture hand movements and cards.}
    \label{fig:recon_examples_app_7}
\end{figure*}

\begin{figure*}
    \centering
    \includegraphics[width=0.94\linewidth]{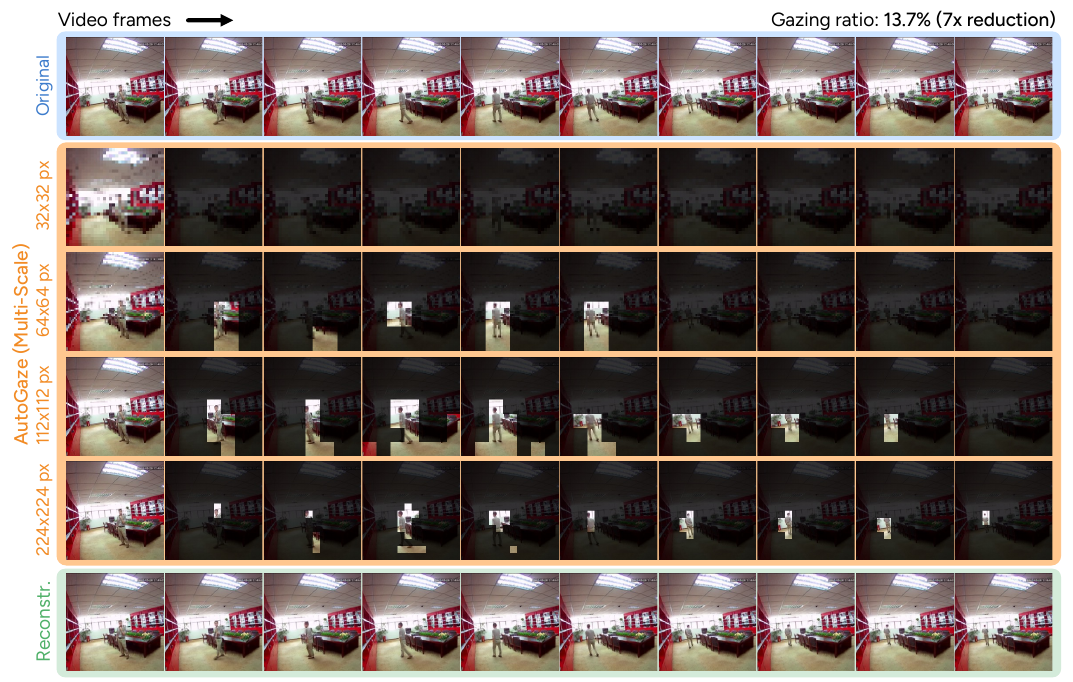}
    \caption{\textbf{Fisheye lens.} \model can select patches that track moving objects with appropriate scales even with lens distortion.}
    \label{fig:recon_examples_app_ood_1}
\end{figure*}

\begin{figure*}
    \centering
    \includegraphics[width=0.94\linewidth]{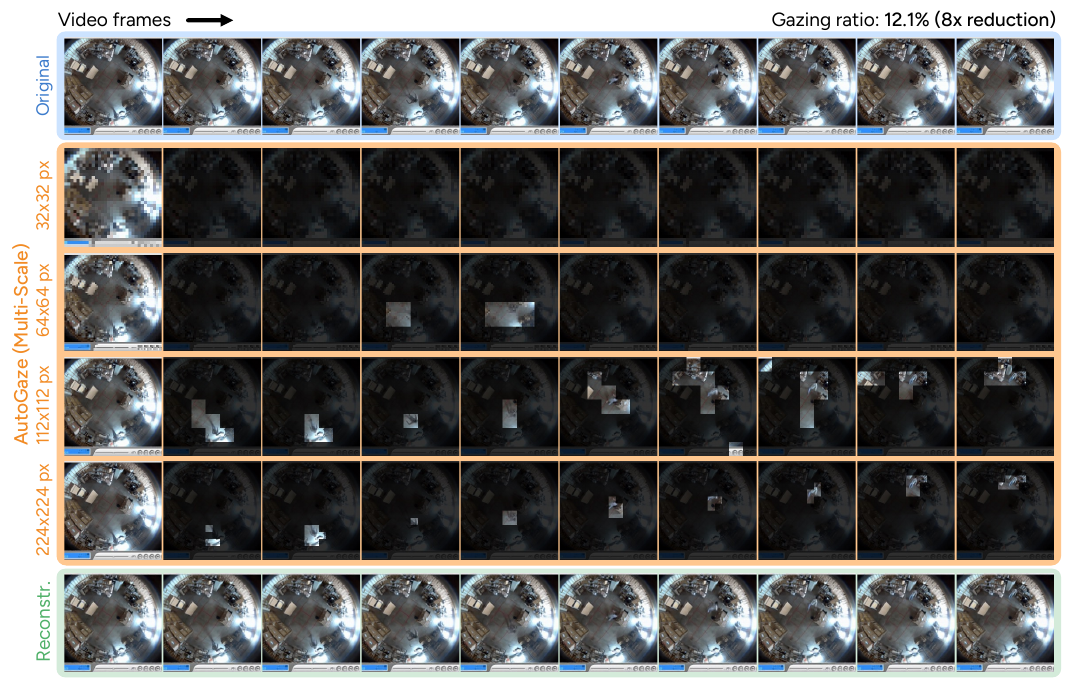}
    \caption{\textbf{Fisheye lens.} In this overhead video with lens distortion, \model tracks the walking person and ignores the static warehouse.}
    \label{fig:recon_examples_app_ood_5}
\end{figure*}

\begin{figure*}
    \centering
    \includegraphics[width=0.94\linewidth]{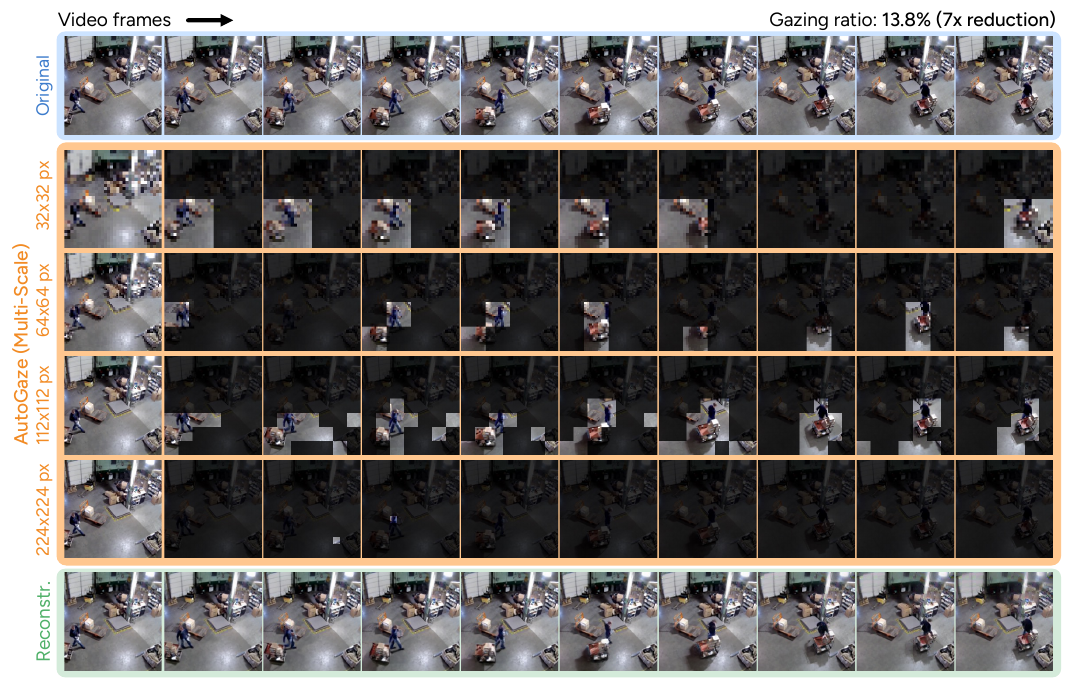}
    \caption{\textbf{Warehouse example.} In this video, \model selects just enough patches to reconstruct the moving person and cart.}
    \label{fig:recon_appendix_warehouse}
\end{figure*}

\begin{figure*}
    \centering
    \includegraphics[width=0.94\linewidth]{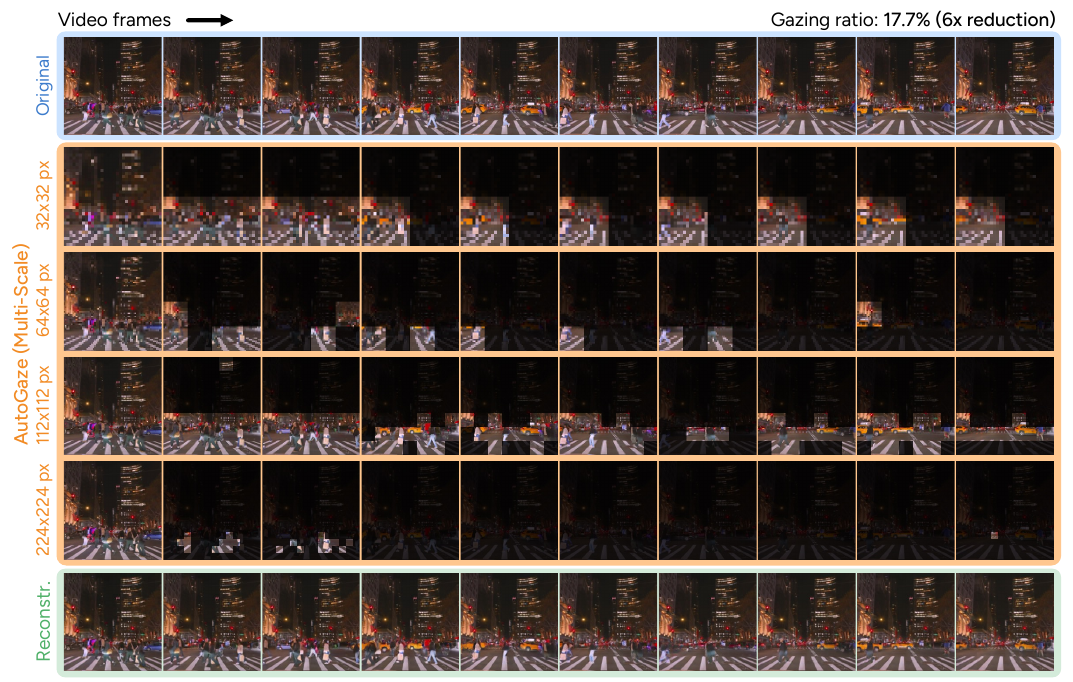}
    \caption{\textbf{Nighttime driving.} \model can be used on nighttime videos such as this one, capturing pedestrians and passing cars.}
    \label{fig:recon_appendix_nightdrive}
\end{figure*}

\begin{figure*}
    \centering
    \includegraphics[width=0.94\linewidth]{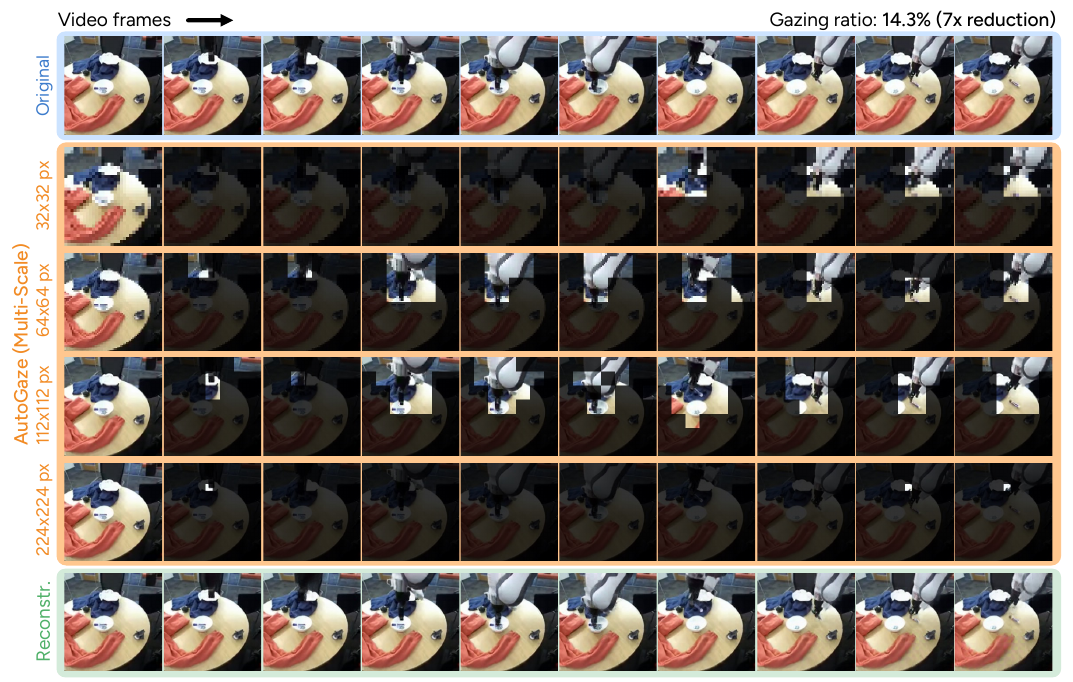}
    \caption{\textbf{Robot arm video.} \model uses different scales to gaze at the robot arm and marker.}
    \label{fig:recon_examples_app_ood_3}
\end{figure*}

\begin{figure*}
    \centering
    \includegraphics[width=0.94\linewidth]{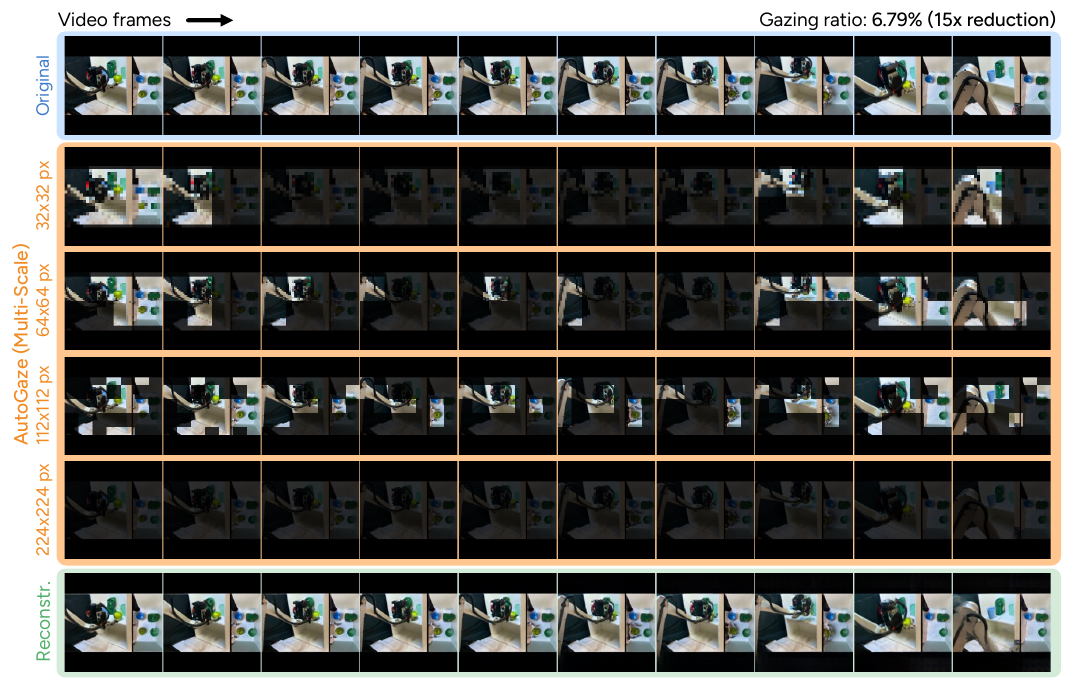}
    \caption{\textbf{Multiple perspectives. } Given two side-by-side videos, \model selects patches in both halves to reconstruct the video.}
    \label{fig:recon_examples_app_ood_2}
\end{figure*}

\begin{figure*}
    \centering
    \includegraphics[width=0.94\linewidth]{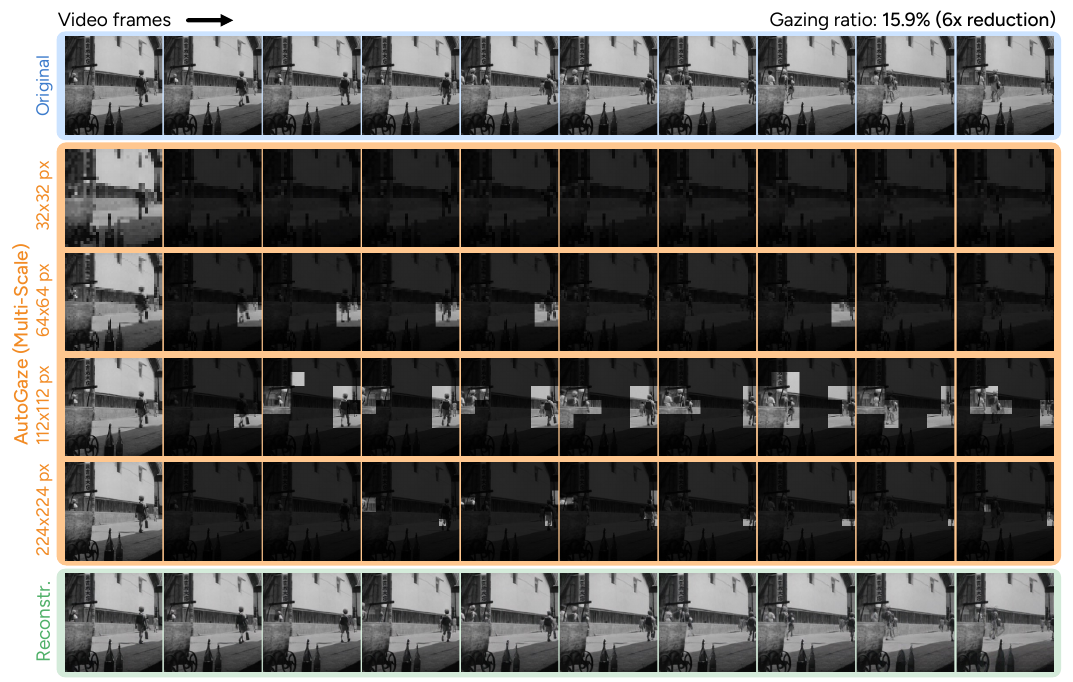}
    \caption{\textbf{Black-and-white film.} In this clip where people are walking, \model uses finer scales to select people as they get smaller.}
    \label{fig:recon_examples_app_ood_4}
\end{figure*}

\begin{figure*}
    \centering
    \includegraphics[width=0.94\linewidth]{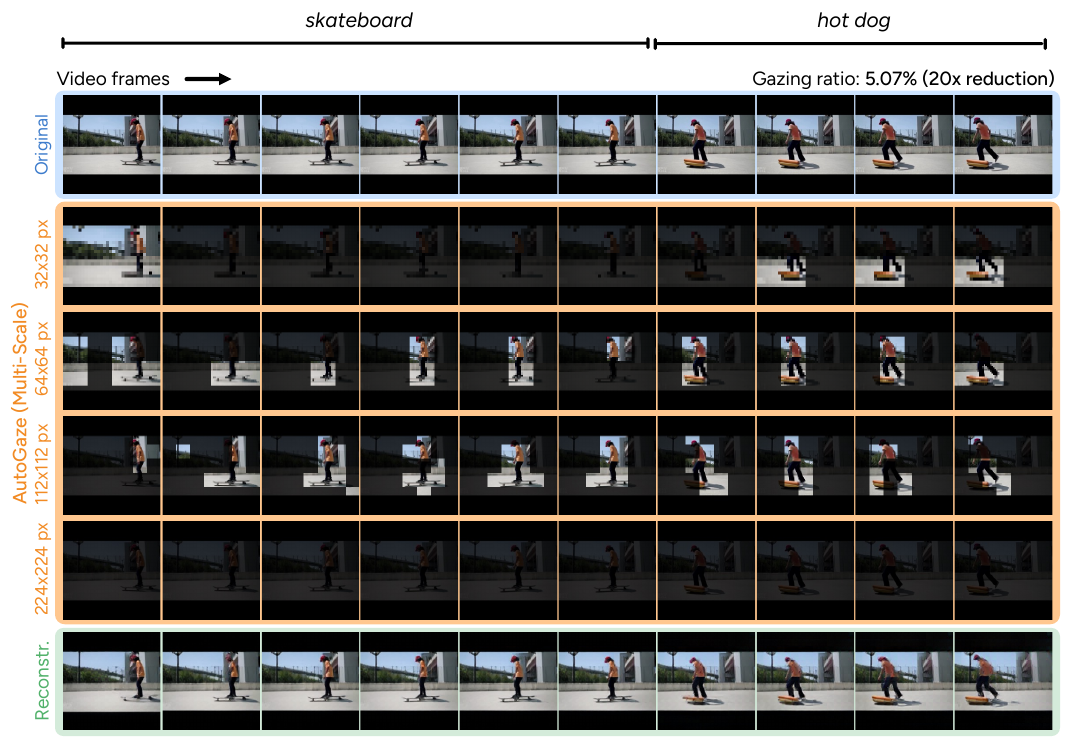}
    \caption{\textbf{Swapping object.} In an OOD scenario where the skateboard swaps out halfway through the video, \model continues to track the moving parts and properly reconstruct each video frame.}
    \label{fig:swap_appendix}
\end{figure*}

Figs \ref{fig:recon_examples_app_1} - \ref{fig:recon_examples_app_ood_4} showcase examples of \model applied to various video domains and OOD use cases, including lectures, sports live stream, cartoons, film clips, picture-in-picture videos, fisheye lens security footage, warehouse surveillance camera, nighttime driving, black-and-white movies, robot arm demonstrations, and split-view videos. Finally, we provide another example of \model continuing to track moving objects even when an object is swapped in the middle of the video (Fig. \ref{fig:swap_appendix}). Refer to each figure caption for particular capabilities demonstrated in that example.

\section{Limitations}
\label{appendix_sec:limitation}
\begin{figure*}
    \centering
    \includegraphics[width=0.94\linewidth]{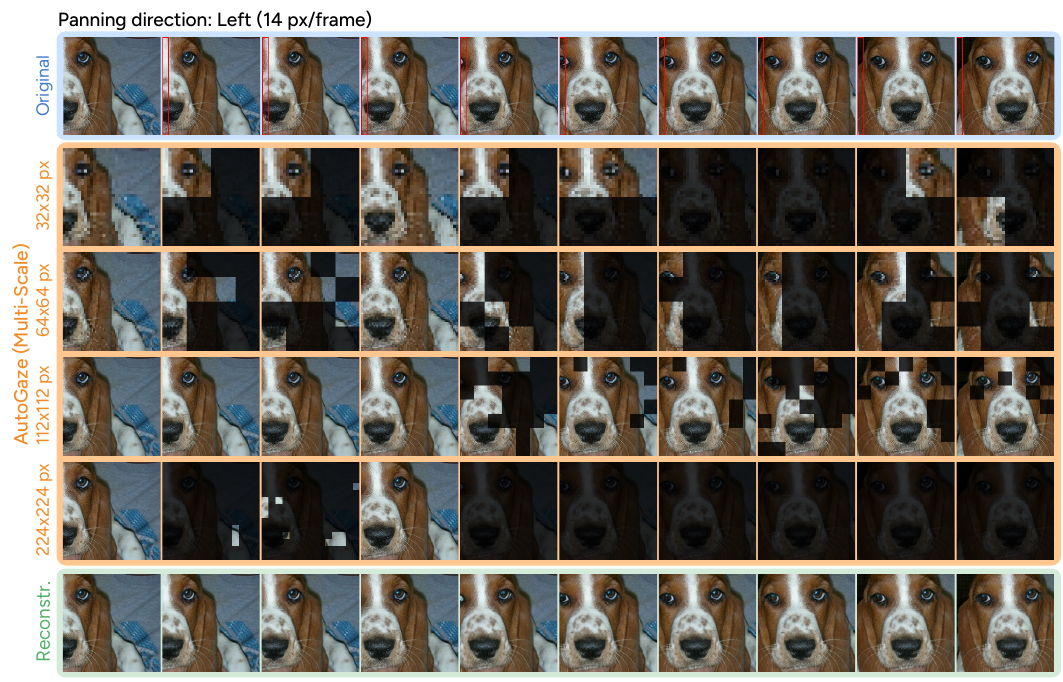}
    \caption{\textbf{Panning over a static image.} As the image slides to the left (new pixels highlighted with a red border), \model does not perfectly track movement; it can select regions that were in different parts of previous frames.}
    \label{fig:no_camera_motion}
\end{figure*}

\begin{figure*}
    \centering
    \includegraphics[width=0.94\linewidth]{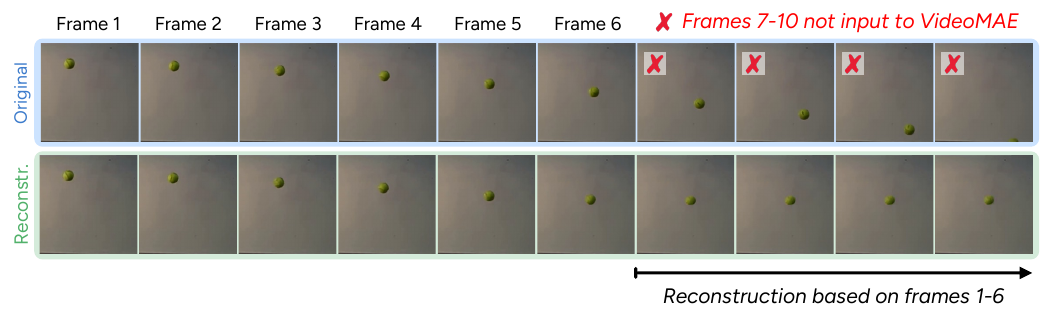}
    \caption{\textbf{Predicting next frames from physics knowledge.} Given the first 6 frames of a ball falling along a parabolic path, VideoMAE is unable to reconstruct the next frames to reflect the continued falling motion.}
    \label{fig:no_physics}
\end{figure*}

We identify two main limitations with \model. First, it does not account for most camera motion; as a scene pans in some direction, it will still subsample patches but not necessarily ignore patches that are redundant up to a shift. See Fig. \ref{fig:no_camera_motion} for an example of this limitation. Second, our model cannot anticipate future frames according to knowledge of physics; though our VideoMAE is causal, it is not trained to have ``intuitive physics'' knowledge (e.g., knowledge that a falling ball will keep falling in the next frame). We illustrate this limitation in Fig. \ref{fig:no_physics} by providing several full video frames of a ball in free fall, and visualizing the VideoMAE's reconstruction of subsequent frames.



